\documentclass[10pt,twocolumn,letterpaper]{article}

\newif\ifarxiv
\arxivtrue  

\ifarxiv
\usepackage[pagenumbers]{cvpr} 
\else
\usepackage{cvpr}              
\fi

\usepackage{microtype}
\usepackage{graphicx}
\usepackage{amsmath}
\usepackage{amssymb}
\usepackage{booktabs}
\usepackage[dvipsnames]{xcolor}
\usepackage{multirow}
\usepackage{stackengine}
\usepackage{paralist}
\usepackage[autolanguage]{numprint}
\usepackage{array}
\newcolumntype{H}{>{\setbox0=\hbox\bgroup}c<{\egroup}@{}}
\usepackage{soul}
\usepackage{cuted}
\usepackage{capt-of}
\usepackage[pagebackref,breaklinks,colorlinks]{hyperref}
\usepackage{balance}
\usepackage{wrapfig}

\usepackage[capitalize]{cleveref}
\crefname{section}{Sec.}{Secs.}
\Crefname{section}{Section}{Sections}
\Crefname{table}{Table}{Tables}
\crefname{table}{Tab.}{Tabs.}

\newcommand\blfootnote[1]{%
  \begingroup
  \renewcommand\thefootnote{}\footnote{#1}%
  \addtocounter{footnote}{-1}%
  \endgroup
}

\newcommand{\todo}[1]{{\color{red} TODO: #1 }}
\newcommand{\ours}[0]{Dream Fields}
\newcommand{\ourssingular}[0]{Dream Field}

\newcommand{\Lclip}[0]{\mathcal{L}_\mathrm{CLIP}}
\newcommand{\pose}[0]{\mathbf{p}}
\newcommand{\image}[0]{\mathbf{I}}
\newcommand{\Ltr}[0]{\mathcal{L}_{T}}
\newcommand{\Ltotal}[0]{\mathcal{L}_{\text{total}}}

\newcommand{\metric}[2]{{\nprounddigits{3}\numprint{#1}$\pm$\small{\numprint{#2}}}}
\newcommand{\pct}[2]{{\nprounddigits{1}\numprint{#1}$\pm$\small{\numprint{#2}}}}
\newcommand{\pctm}[2]{{\nprounddigits{1}\numprint{#1}}}  
\newcommand{\lituu}[0]{{LiT$_\text{uu}$}}

\definecolor{darkgreen}{rgb}{0.3, 0.7, 0.3}

\def\cvprPaperID{11674}
\def\confName{CVPR}
\def\confYear{2022}

\begin{document}

\title{Zero-Shot Text-Guided Object Generation with \ours{}}

\author{Ajay Jain$^{1,2*}$
\and Ben Mildenhall$^2$
\and Jonathan T. Barron$^2$
\and Pieter Abbeel$^1$
\and Ben Poole$^2$}
\maketitle

\begin{strip}%
    \centering
    \begin{minipage}[t]{0.55\linewidth}
    \ifarxiv
        \includegraphics[trim={0 21cm 30cm 0},clip,width=\linewidth]{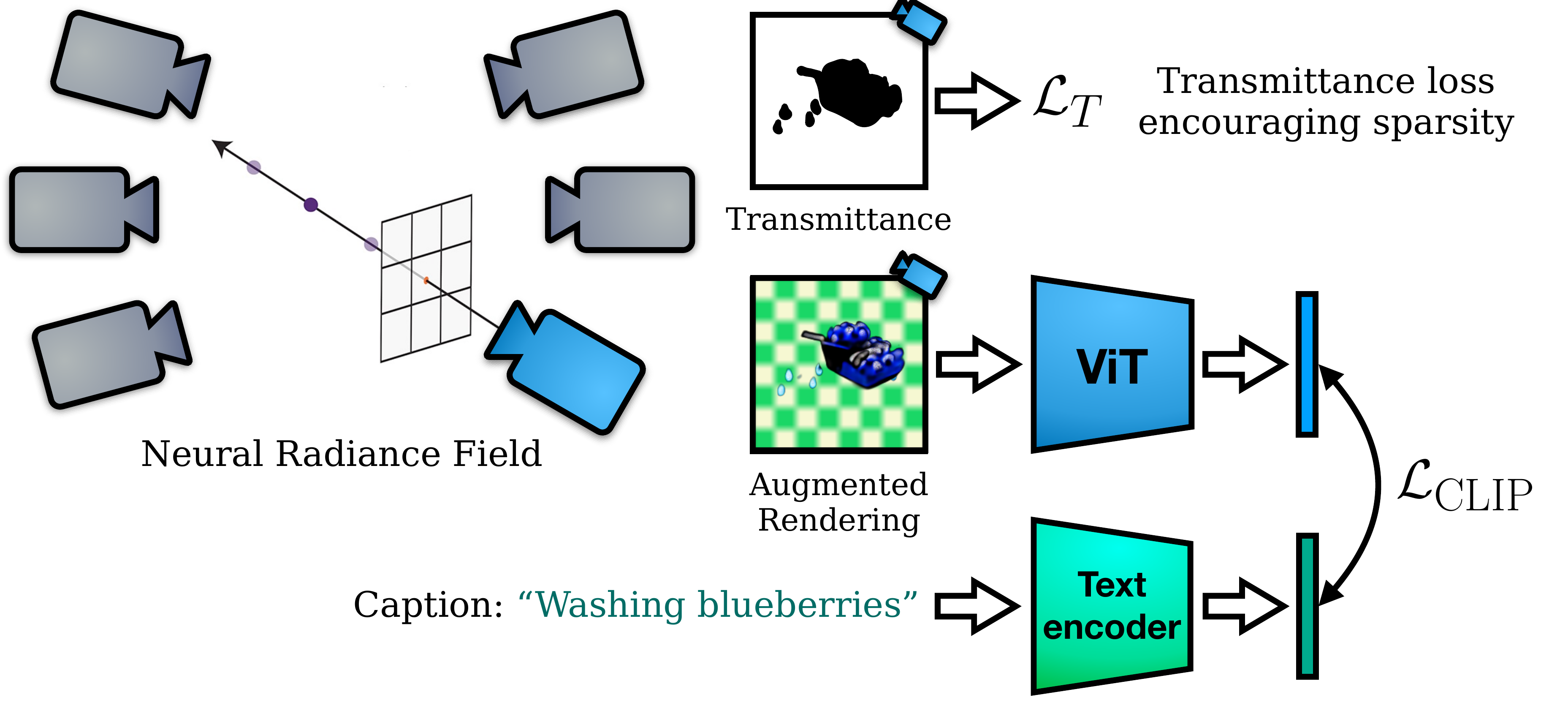}%
    \else
        \includegraphics[trim={0 21cm 30cm 0},clip,width=\linewidth]{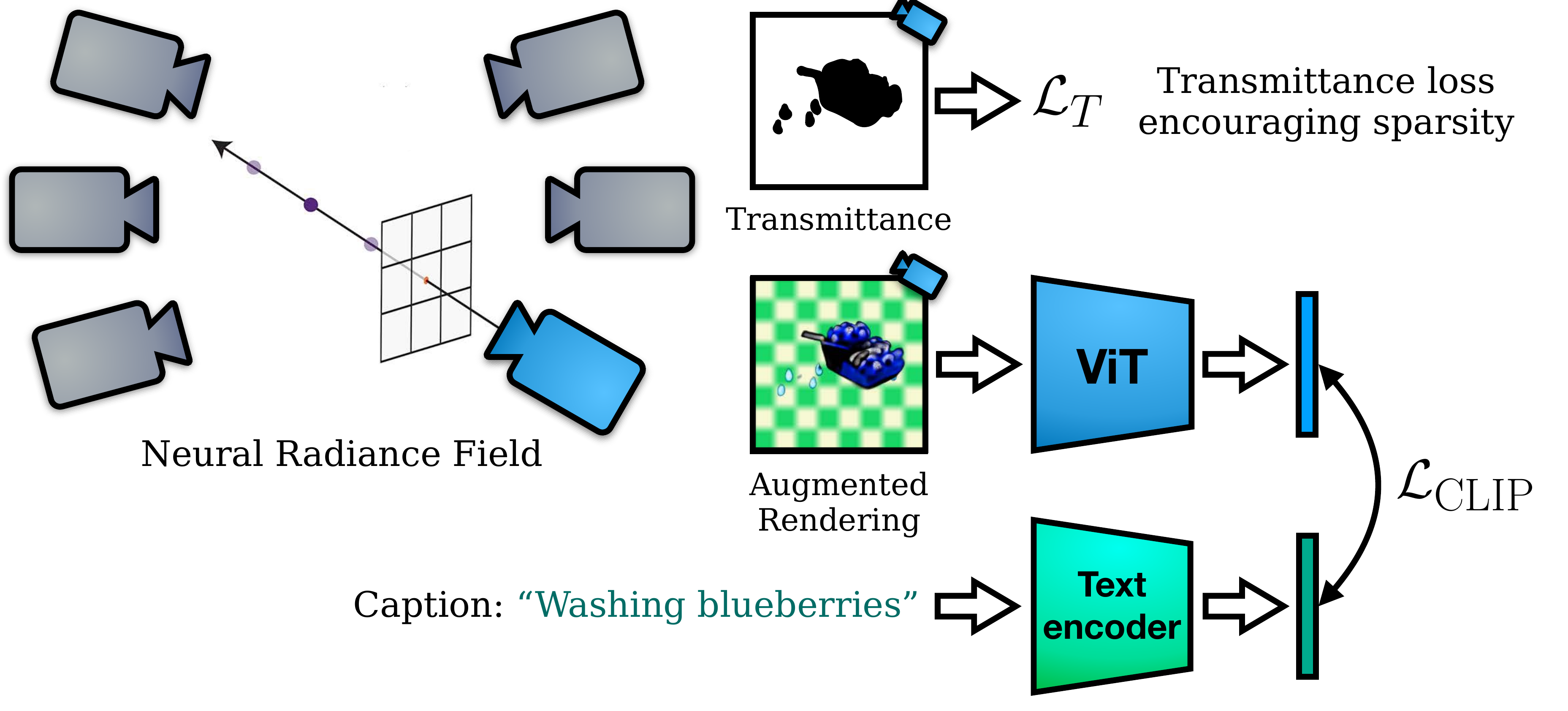}%
    \fi
    \captionof{figure}{Given a caption, we learn a {\textbf{\ourssingular{}}}, a continuous volumetric representation of an object's geometry and appearance learned with guidance from a pre-trained model. We optimize the \ourssingular{} by rendering images of the object from random camera poses that are scored with {\em frozen} pre-trained {\color{blue} \textbf{image}} and {\color{darkgreen} \textbf{text}} encoders trained on web images and alt-text. 2D views share the same underlying radiance field for consistent geometry.}
    \label{fig:teaser:method}%
    \end{minipage}%
    \hspace{2mm}
    \begin{minipage}[t]{0.4\linewidth}%
    \ifarxiv
    \includegraphics[trim={8cm 35cm 8cm 0},clip,width=\linewidth]{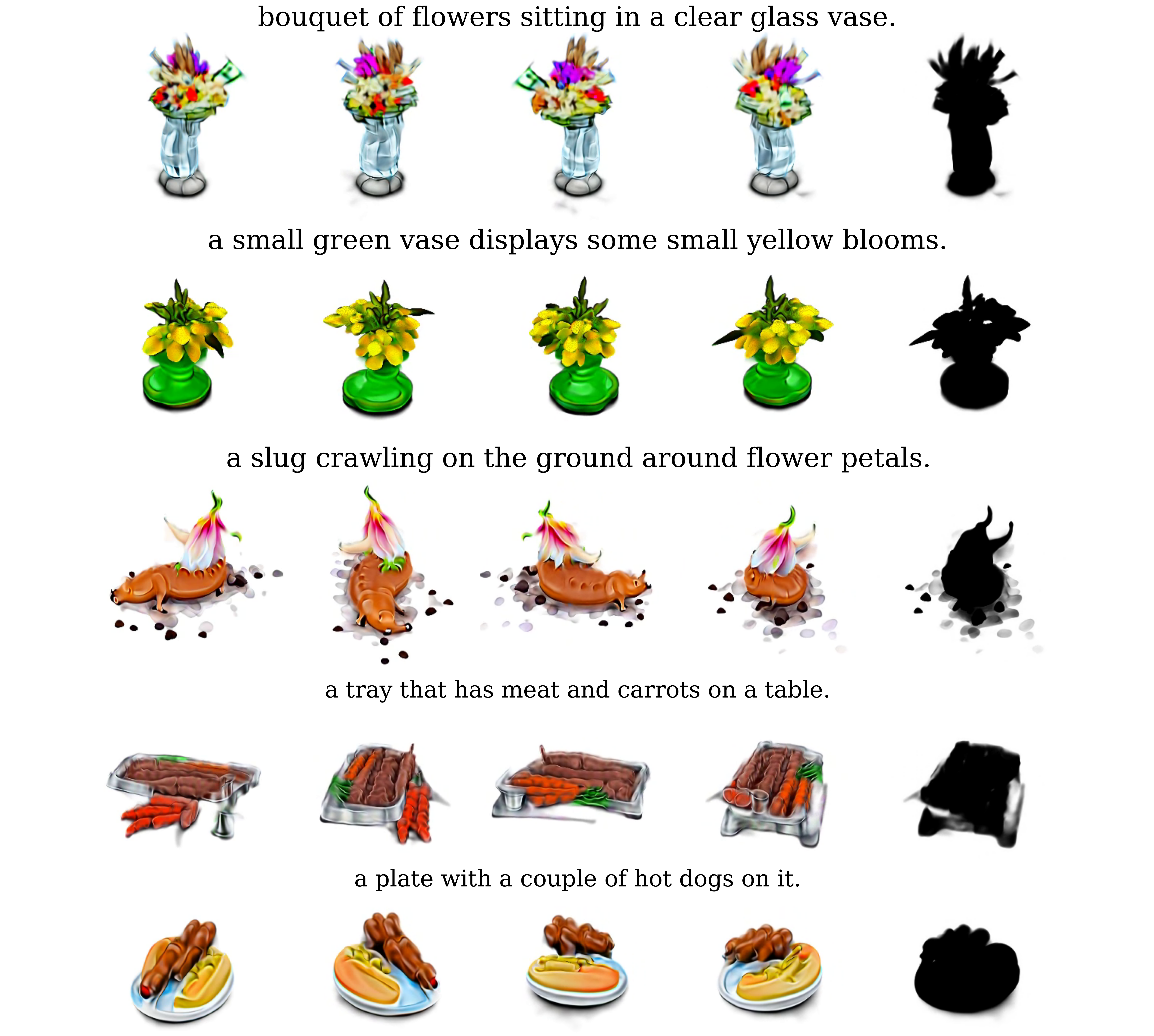}%
    \else
    \includegraphics[width=\linewidth]{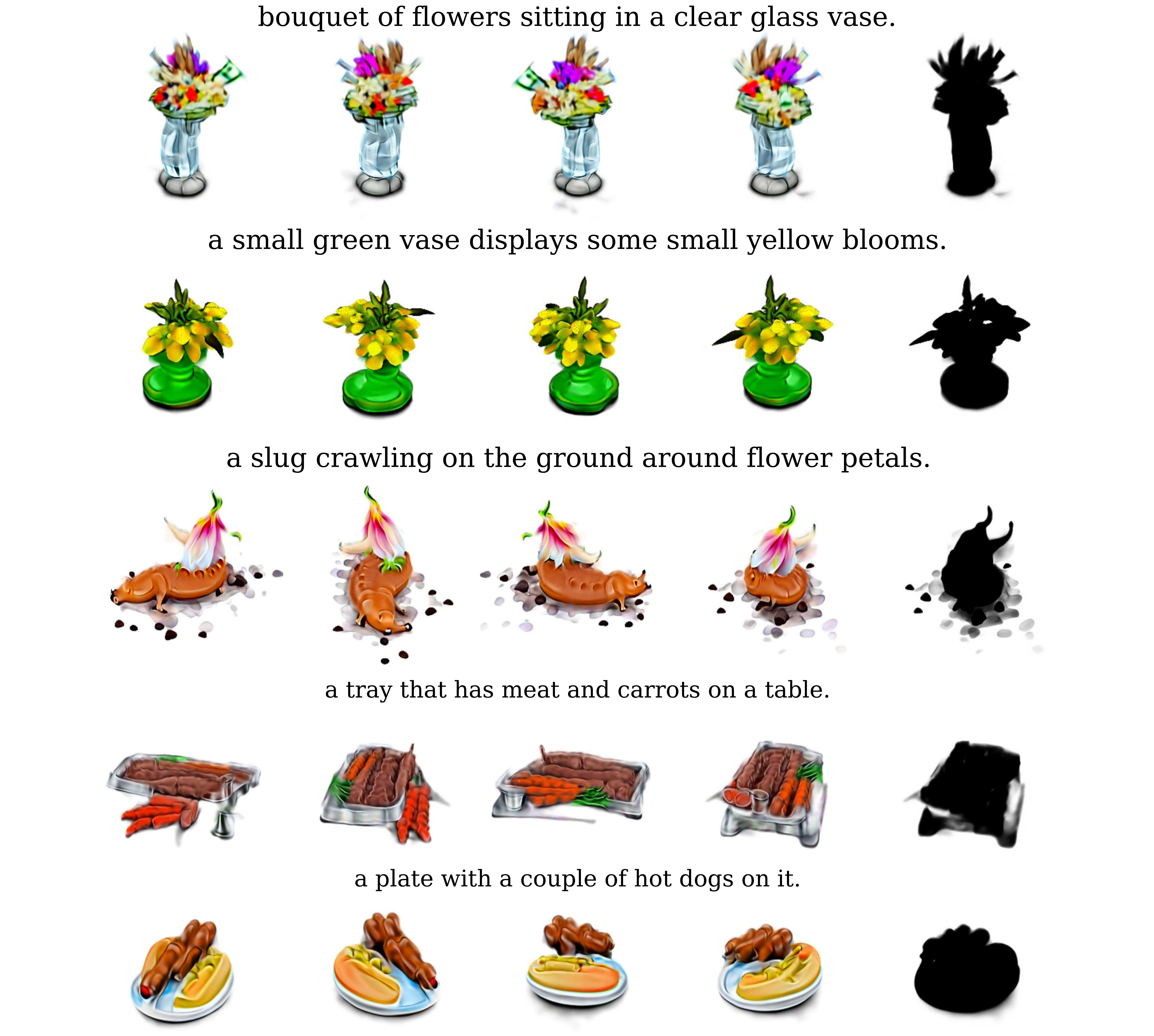}%
    \fi
    \captionof{figure}{Example \ours{} rendered from four perspectives. On the right, we show transmittance from the final perspective. We create diverse outputs using the compositionality of language; these captions from MSCOCO describe three flower arrangements with different properties like context and color.}%
    \label{fig:teaser:qual}%
    \end{minipage}
\end{strip}

\begin{abstract}
We combine neural rendering with multi-modal image and text representations to synthesize diverse 3D objects solely from natural language descriptions. Our method, \ours{}, can generate the geometry and color of a wide range of objects without 3D supervision. Due to the scarcity of diverse, captioned 3D data, prior methods only generate objects from a handful of categories, such as ShapeNet. Instead, we guide generation with image-text models pre-trained on large datasets of captioned images from the web. Our method optimizes a Neural Radiance Field from many camera views so that rendered images score highly with a target caption according to a pre-trained CLIP model. To improve fidelity and visual quality, we introduce simple geometric priors, including sparsity-inducing transmittance regularization, scene bounds, and new MLP architectures. In experiments, \ours{} produce realistic, multi-view consistent object geometry and color from a variety of natural language captions. \blfootnote{$^1$UC Berkeley, $^2$Google Research.} \blfootnote{$^*$Work done at Google.} \blfootnote{Correspondence to {\small \texttt{ajayj@berkeley.edu}}.} \blfootnote{Project website and code: {\url{https://ajayj.com/dreamfields}}}
\end{abstract}

\vspace{-1cm}
\section{Introduction}
\label{sec:intro}

Detailed 3D object models bring multimedia experiences to life. Games, virtual reality applications and films are each populated with thousands of object models, each designed and textured by hand with digital software.
While expert artists can author high-fidelity assets, the process is painstakingly slow and expensive.
Prior work leverages 3D datasets to synthesize shapes in the form of point clouds, voxel grids, triangle meshes, and implicit functions using generative models like GANs~\cite{3dgan, shapegf, gupta2020neural, Zhou_2021_ICCV}. These approaches only support a few object categories due to small labeled 3D shape datasets. But multimedia applications require a wide variety of content, and need both 3D geometry and texture.

In this work, we propose \ours{}, a method to automatically generate open-set 3D models from natural language prompts. Unlike prior work, our method does not require any 3D training data, and uses natural language prompts that are easy to author with an expressive interface for specifying desired object properties. We demonstrate that the compositionality of language allows for flexible creative control over shapes, colors and styles.

A \ourssingular{} is a Neural Radiance Field (NeRF) trained to maximize a deep perceptual metric with respect to both the geometry and color of a scene. NeRF and other neural 3D representations have recently been successfully applied to novel view synthesis tasks where ground-truth RGB photos are available. NeRF is trained to reconstruct images from multiple viewpoints. As the learned radiance field is shared across viewpoints, NeRF can interpolate between viewpoints smoothly and consistently. Due to its neural representation, NeRF can be sampled at high spatial resolutions unlike voxel representations and point clouds, and are easy to optimize unlike explicit geometric representations like meshes as it is topology-free.

However, existing photographs are not available when creating novel objects from descriptions alone. Instead of learning to reconstruct known input photos, we learn a radiance field such that its renderings have high semantic similarity with a given text prompt. We extract these semantics with pre-trained neural image-text retrieval models like CLIP \cite{clip:DBLP:journals/corr/abs-2103-00020}, learned from hundreds of millions of captioned images. As NeRF's volumetric rendering and CLIP's image-text representations are differentiable, we can optimize \ours{} end-to-end for each prompt. Figure~\ref{fig:teaser:method} illustrates our method.

\begin{figure}[t]
  \centering
  \ifarxiv
    \includegraphics[width=\linewidth]{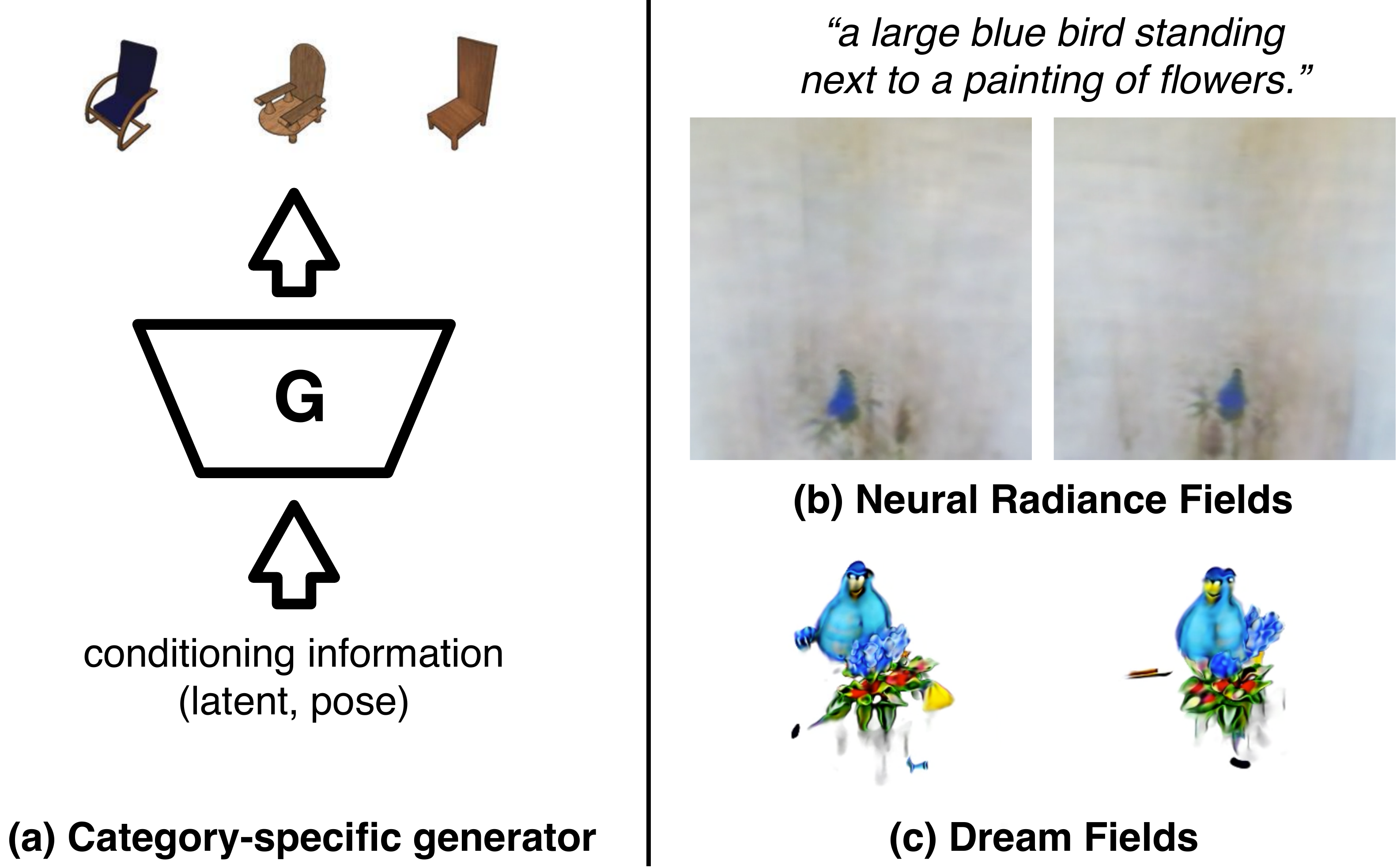}
  \else
    \includegraphics[width=\linewidth]{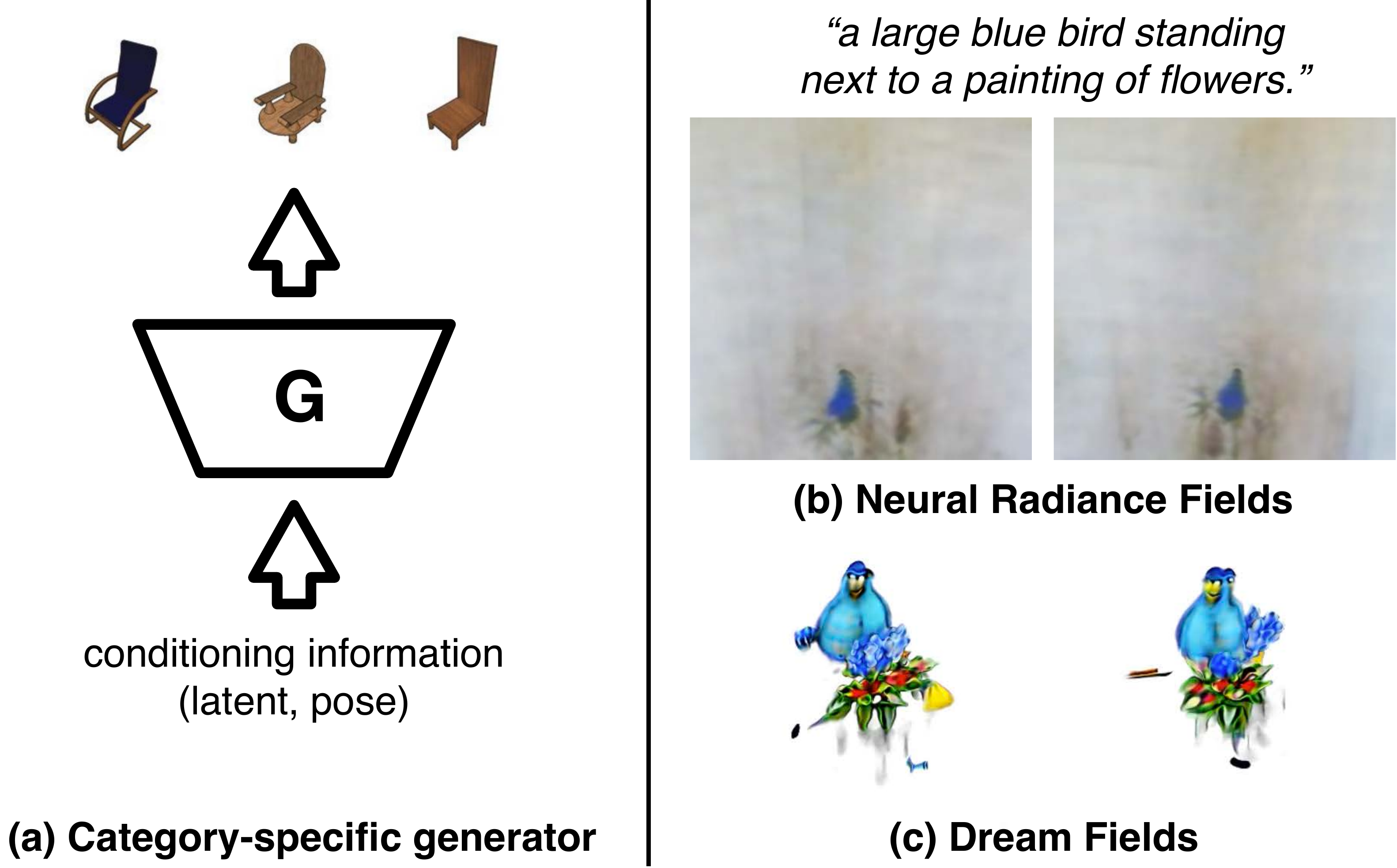}
  \fi
    \caption{\textbf{Challenges of text-to-3D synthesis:} \textbf{(a) Poor generalization from limited 3D datasets:} Most 3D generative models are learned on datasets of specific object categories like ShapeNet~\cite{shapenet2015}, and won't generalize to novel concepts zero-shot. \textbf{(b) Neural Radiance Fields are too flexible without multi-view supervision:} NeRF learns to represent geometry and texture from scene-specific multi-view data, so it does not require a diverse dataset of objects. Yet, when only a source caption is available instead of multi-view images, NeRF produces significant artifacts (\eg, near field occlusions). \textbf{(c) \ours{}:} We introduce general geometric priors that retain much of NeRF's flexibility while improving realism.}
    \label{fig:challenges}
\end{figure}

In experiments, \ours{} learn significant artifacts if we naively optimize the NeRF scene representation with textual supervision without adding additional geometric constraints~(Figure~\ref{fig:challenges}). We propose general-purpose priors and demonstrate that they greatly improve the realism of results. Finally, we quantitatively evaluate open-set generation performance using a dataset of diverse object-centric prompts.

\noindent{}Our contributions include:
\begin{compactitem}
    \item Using aligned image and text models to optimize NeRF without 3D shape or multi-view data,
    \item \ours{}, a simple, constrained 3D representation with neural guidance that supports diverse 3D object generation from captions in zero-shot, and
    \item Simple geometric priors including transmittance regularization, scene bounds, and an MLP architecture that together improve fidelity.
\end{compactitem}

\section{Related Work}
\label{sec:related}
Our work is primarily inspired by DeepDream \cite{mordvintsev2015inceptionism} and other methods for visualizing the preferred inputs and features of neural networks by optimizing in image space \cite{olah2017feature,nguyen2016synthesizing, nguyen2017plug}. These methods enable the generation of interesting images from a pre-trained neural network without the additional training of a generative model. Closest to our work is \cite{mordvintsev2018differentiable}, which studies differentiable image parameterizations in the context of style transfer. Our work replaces the style and content-based losses from that era with an image-text loss enabled by progress in contrastive representation learning on image-text datasets \cite{desai2021virtex, clip:DBLP:journals/corr/abs-2103-00020,pmlr-v139-jia21b,zhai2021lit}.
The use of image-text models enables easy and flexible control over the style and content of generated imagery through textual prompt design. We optimize both geometry and color using the differentiable volumetric rendering and scene representation provided by NeRF, whereas \cite{mordvintsev2018differentiable} was restricted to fixed geometry and only optimized texture. Together these advances enable a fundamentally new capability: open-ended text-guided generation of object geometry and texture.

Concurrently to \ours{}, a few early works have used CLIP \cite{clip:DBLP:journals/corr/abs-2103-00020} to synthesize or manipulate 3D object representations. CLIP-Forge \cite{sanghi2021clipforge} generates multiple object geometries from text prompts using a CLIP embedding-conditioned normalizing flow model and geometry-only decoder trained on ShapeNet categories. Still, CLIP-Forge generalizes poorly outside of ShapeNet categories and requires ground-truth multi-view images and voxel data. 
Text2Shape \cite{text2shape:DBLP:journals/corr/abs-1803-08495} learns a text-conditional Wasserstein GAN \cite{pmlr-v70-arjovsky17a, NIPS2014_5ca3e9b1} to synthesize novel voxelized objects, but only supports finite resolution generation of individual ShapeNet categories. 
In \cite{chu2021evolving}, object geometry is optimized evolutionarily for high CLIP score from a single view then manually colored. ClipMatrix \cite{jetchev2021clipmatrix} edits the vertices and textures of human SMPL models \cite{SMPL:2015} to create stylized, deformable humanoid meshes. 
\cite{perez2021sculpting} creates an interactive interface to edit signed-distance fields in localized regions, though they do not optimize texture or synthesize new shapes. Text-based manipulation of existing objects is complementary to us.

For images, there has been an explosion of work that leverages CLIP to guide image generation. Digital artist Ryan Murdock ({\tt\footnotesize \href{https://twitter.com/advadnoun}{@advadnoun}}) used CLIP to guide learning of the weights of a SIREN network \cite{sitzmann2020implicit}, similar to NeRF but without volume rendering and focused on image generation. Katherine Crowson ({\tt\footnotesize \href{https://twitter.com/rivershavewings}{@rivershavewings}}) combined CLIP with optimization of VQ-GAN codes \cite{esser2021taming} and used diffusion models as an image prior \cite{dhariwal2021diffusion}. Recent work from Mario Klingemann ({\tt\footnotesize \href{https://twitter.com/quasimondo}{@quasimondo}}) and \cite{patashnik2021styleclip} have shown how CLIP can be used to guide GAN models like StyleGAN \cite{karras2020analyzing}. Some works have optimized parameters of vector graphics, suggesting CLIP guidance is highly general~\cite{jain21vector,frans2021clipdraw,schaldenbrand2021styleclipdraw}. These methods highlighted the surprising capacity of what image-text models have learned and their utility for guiding 2D generative processes.  Direct text to image synthesis with generative models has also improved tremendously in recent years \cite{dalle:DBLP:journals/corr/abs-2102-12092, zhang2021cross}, but requires training large generative models on large-scale datasets, making such methods challenging to directly apply to text to 3D where no such datasets exist.

There is also growing progress on generative models with NeRF-based generators trained solely from 2D imagery. However, these models are category-specific and trained on large datasets of mostly forward-facing scenes~\cite{chanmonteiro2020pi-GAN, Niemeyer2020GIRAFFE, schwarz2020graf, gu2021stylenerf, zhou2021CIPS3D}, lacking the flexibility of open-set text-conditional models. Shape-agnostic priors have been used for 3D reconstruction~\cite{BarronTPAMI2015, shapehd, genre}.

\section{Background}
\label{sec:background}
Our method combines Neural Radiance Fields (NeRF) \cite{mildenhall2020nerf} with an image-text loss from \cite{clip:DBLP:journals/corr/abs-2103-00020}. We begin by discussing these existing methods, and then detail our improved approach and methodology that enables high quality text to object generation.
\subsection{Neural Radiance Fields}
\label{sec:background:nerf}

NeRF \cite{mildenhall2020nerf} parameterizes a scene's density and color using a multi-layer perceptron (MLP) with parameters $\theta$ trained with a photometric loss relying on multi-view photographs of a scene. 
In our simplified model, the NeRF network takes in a 3D position $\mathbf x$ and outputs parameters for an emission-absorption volume rendering model: density $\sigma_\theta({\mathbf x})$ and color $\mathbf c_\theta({\mathbf x})$. 
Images can be rendered from desired viewpoints by integrating color along an appropriate ray, $\mathbf{r}(t)$, for each pixel according to the volume rendering equation:
\begin{gather}
    \mathbf C(r, \theta) = \int_{t_n}^{t_f} T(\mathbf r, t) \sigma_\theta(\mathbf r(t)) \mathbf c_\theta(\mathbf r(t)) dt, \\
    \text{where } T(\mathbf r,\theta, t) = \exp\left( -\int_{t_n}^t \sigma_\theta(\mathbf r(s)) ds\right)   \, .
    \label{eq:transmittance_continuous}
\end{gather}
The integral $T(\mathbf r, \theta, t)$ is known as ``transmittance'' and describes the probability that light along the ray will not be absorbed when traveling from $t_n$ (the near scene bound) to $t$. In practice~\cite{mildenhall2020nerf}, these two integrals are approximated by breaking up the ray into smaller segments $[t_{i-1}, t_i)$ within which $\sigma$ and $\mathbf c$ are assumed to be roughly constant:
\begin{gather}
    \mathbf C(\mathbf r, \theta) \approx \sum_i T_i (1-\exp(-\sigma_\theta(\mathbf r(t_i))\delta_i)) \mathbf c_\theta(\mathbf r(t_i)) \\
    T_i = \exp\left(- \textstyle\sum_{j < i} \sigma_\theta(\mathbf r(t_j))\delta_j \right)  , \quad \delta_i = t_i - t_{i-1} \, .
    \label{eq:transmittance_discrete}
\end{gather}
For a given setting of MLP parameters $\theta$ and pose $\pose{}$, we determine the appropriate ray for each pixel, compute rendered colors $C(\mathbf r, \theta)$ and transmittances, and gather the results to form the rendered image, $I(\theta, \pose{})$ and transmittance $T(\theta, \pose{})$.

In order for the MLP to learn high frequency details more quickly~\cite{tancik2020fourfeat}, the input $\mathbf x$ is preprocessed by a sinusoidal positional encoding $\gamma$ before being passed into the network:
\begin{align}
    \gamma(\mathbf x) = \left[ \cos(2^l \mathbf x), \sin(2^l \mathbf x)  \right]_{l=0}^{L-1} \, ,
\end{align}
where $L$ is referred to as the number of ``levels'' of positional encoding. In our implementation, we specifically apply the integrated positional encoding (IPE) proposed in mip-NeRF to combat aliasing artifacts~\cite{barron2021mipnerf} combined with a random Fourier positional encoding basis~\cite{tancik2020fourfeat} with frequency components sampled according to
\begin{align}
    \mathbf \omega = 2^u \mathbf d, \quad \text{where } u\sim \mathcal{U}[0, L], \,\, \mathbf d \sim \mathcal{U}(\mathbb S^2) \, .
\end{align}

\subsection{Image-text models}
\label{sec:background:clip}

Large-scale datasets of images paired with associated text have enabled training large-scale models that can accurately score whether an image and an associated caption are likely to correspond \cite{clip:DBLP:journals/corr/abs-2103-00020, pmlr-v139-jia21b, desai2021virtex}. 
These models consist of an image encoder $\mathbf g$, and text encoder $\mathbf h$, that map images and text into a shared embedding space. Given a sentence $y$ and an image $\image{}$, these image-text models produce a scalar score: $\mathbf g(\image{})^\mathrm{T} \mathbf h(\mathbf y)$ that is high when the text is a good description of the image, and low when the the image and text are mismatched. Note that the embeddings $\mathbf g(\image{})$ and $\mathbf h(\mathbf y)$ are often normalized, i.e. $\|\mathbf g(\image{})\| = \|\mathbf h(\mathbf y)\| = 1$. Training is typically performed with a symmetric version of the InfoNCE loss \cite{oord2018representation, poole2019variational} that aims to maximize a variational lower bound on the mutual information between images and text. Prior work has shown that once trained, the image and text encoders are useful for a number of downstream tasks \cite{clip:DBLP:journals/corr/abs-2103-00020, zhai2021lit}. In \cite{dalle:DBLP:journals/corr/abs-2102-12092}, the image and text encoders are used to score the correspondence of outputs of a generative image model to a target caption \cite{dalle:DBLP:journals/corr/abs-2102-12092}. We build on this work by optimizing a volume to produce a high-scoring image, not just reranking.

\section{Method}
\label{sec:method}

In this section, we develop \ours{}: a zero-shot object synthesis method given only a natural language caption.

\subsection{Object representation}

Building on the NeRF scene representation~(Section~\ref{sec:background:nerf}), a \ourssingular{} optimizes an MLP with parameters $\theta$ that produces outputs $\sigma_\theta(\mathbf x)$ and $\mathbf{c}_\theta(\mathbf x)$ representing the differential volume density and color of a scene at every 3D point $\mathbf{x}$. This field expresses object geometry via the density network. Our object representation is only dependent on 3D coordinates and not the camera's viewing direction, as we did not find it beneficial. Given a camera pose $\pose{}$, we can render an image $\image{}(\theta, \pose{})$ and compute the transmittance $T(\theta, \pose{})$ using $N$ segments via \eqref{eq:transmittance_discrete}. Segments are spaced at roughly equal intervals with random jittering along the ray. The number of segments, $N$, determines the fidelity of the rendering. In practice, we fix it to 192 during optimization.

\subsection{Objective}
\label{sec:method:objective}
How can we train a \ourssingular{} to represent a given caption? If we assume that an object can be described similarly when observed from any perspective, we can randomly sample poses and try to enforce that the rendered image matches the caption at all poses. We can implement this idea by using a CLIP network to measure the match between a caption and image given parameters $\theta$ and pose $\pose{}$:
\begin{equation}
    \Lclip{}(\theta, \text{pose}~\pose{}, \text{caption}~y) = -\mathbf g(\image{}(\theta, \pose{}))^\mathrm{T} \mathbf h(\mathbf y)
    \label{eq:lclip}
\end{equation}
where $\mathbf g(\cdot)$ and $\mathbf h(\cdot)$ are aligned representations of image and text semantics, and $\image{}(\theta, \pose{})$ is a rendered image of the scene from camera pose $\pose{}$. Each iteration of training, we sample a pose $\pose{}$ from a prior distribution, render $\image{}$, and minimize $\Lclip{}$ with respect to the parameters of the \ourssingular{} MLP, $\theta$. Equation~\eqref{eq:lclip} measures the similarity of an image and the provided caption in feature space.

We primarily use image and text encoders  from CLIP~\cite{clip:DBLP:journals/corr/abs-2103-00020}, which has a Vision Transformer image encoder $\mathbf g(\cdot)$~\cite{dosovitskiy2020vit} and masked transformer text encoder $\mathbf h(\cdot)$~\cite{vaswani2017attention} trained contrastively on a large dataset of 400\textsc{M} captioned 224$^2$ images. We also use a baseline Locked Image-Text Tuning (LiT) ViT B/32 model from~\cite{zhai2021lit} trained via the same procedure as CLIP on a larger dataset of billions of higher-resolution (288$^2$) captioned images. The LiT training set was collected following a simplified version of the ALIGN web alt-text dataset collection process~\cite{pmlr-v139-jia21b} and includes noisy captions.

Figure~\ref{fig:teaser:method} shows a high-level overview of our method.
DietNeRF~\cite{Jain_2021_ICCV} proposed a related semantic consistency regularizer for NeRF based on the idea that ``a bulldozer is a bulldozer from any perspective''. The method computed the similarity of a rendered and a real image. In contrast, \eqref{eq:lclip} compares rendered images and a \emph{caption}, allowing it to be used in zero-shot settings when there are no object photos.

\subsection{Challenges with CLIP guidance}
Due to their flexibility, Neural Radiance Fields are capable of high-fidelity novel view synthesis on a tremendous diversity of real-world scenes when supervised with multi-view consistent images. Their reconstruction loss will typically learn to remove artifacts like spurious density when sufficiently many input images are available. However, we find that the NeRF scene representation is too unconstrained when trained solely with $\Lclip$ \eqref{eq:lclip} alone from a discrete set of viewpoints, resulting in severe artifacts that satisfy $\Lclip$ but are not visually compatible according to humans (see Figure~\ref{fig:challenges}b). 
NeRF learns high-frequency and near-field~\cite{kaizhang2020} artifacts like partially-transparent ``floating`` regions of density. It also fills the entire camera viewport rather than generating individual objects. Geometry is unrealistic, though textures reflect the caption, reminiscent of the artifacts in Deep Dream feature visualizations~\cite{mordvintsev2015inceptionism, olah2017feature}.

\subsection{Pose sampling}

Image data augmentations such as random crops are commonly used to improve and regularize image generation in DeepDream\cite{mordvintsev2015inceptionism} and related work. Image augmentations can only use in-plane 2D transformations. \ours{} support 3D data augmentations by sampling different camera pose extrinsics at each training iteration. We uniformly sample camera azimuth in 360$^\circ$ around the scene, so each training iteration sees a different orientation of the object. As the underlying scene representation is shared, this improves the realism of object geometry. For example, sampling azimuth in a narrow interval tended to create flat, billboard geometry.

The camera elevation, focal length and distance from the subject can also be augmented, but we did not find this necessary. Instead, we use a fixed camera focal length during optimization that is scaled by $m_\mathrm{focal}=1.2$ to enlarge the object 20\%. Rendering cost is constant in the focal length.

\begin{figure}[t]
    \centering
    \ifarxiv
        \includegraphics[width=\linewidth]{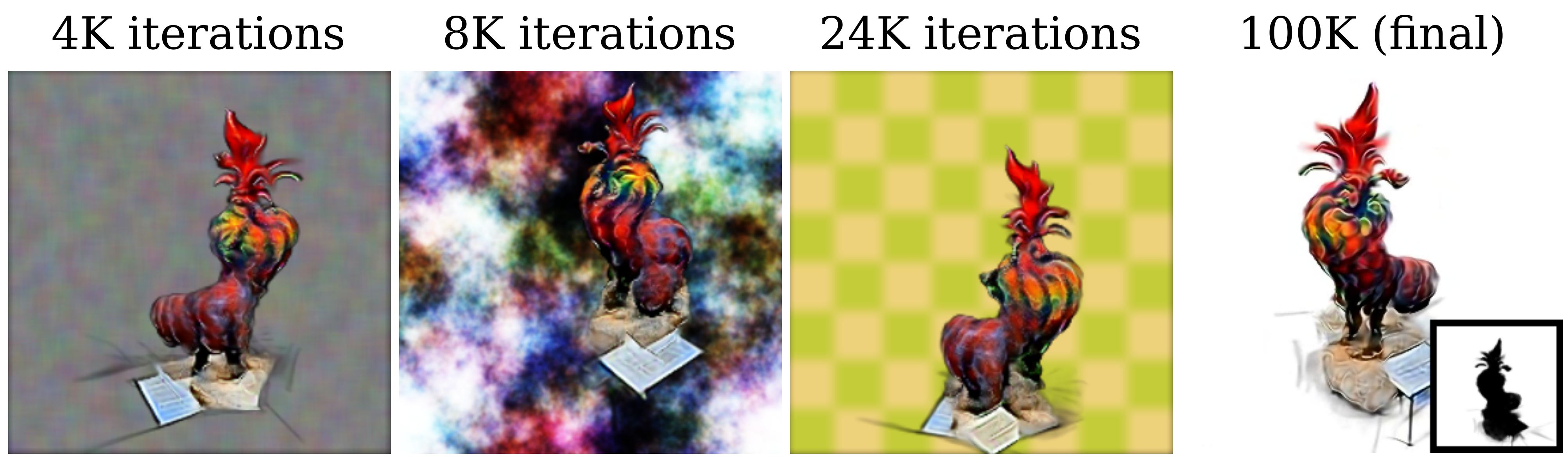}
    \else
        \includegraphics[width=\linewidth]{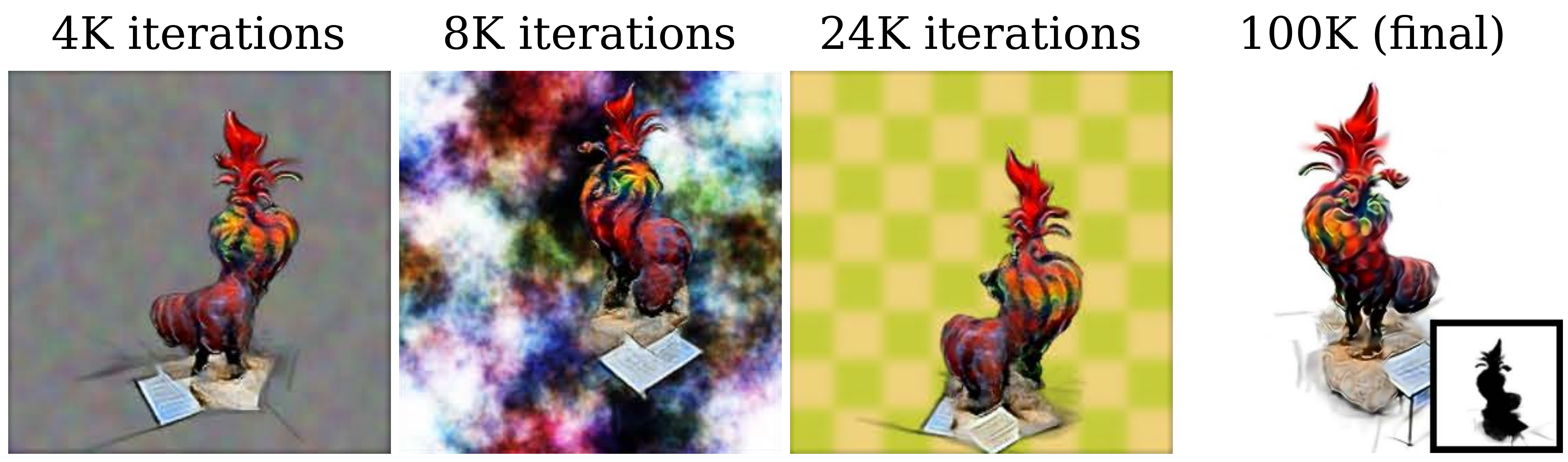}
    \fi
    \caption{To encourage coherent foreground objects, \ours{} train with 3 types of background augmentations: blurred Gaussian noise, textures and checkerboards. At test time, we render with a white background. \textbf{Prompt:} \textit{``A sculpture of a rooster.''}}
    \label{fig:bgaugs}
\end{figure}

\subsection{Encouraging coherent objects through sparsity}

To remove near-field artifacts and spurious density, we regularize the opacity of \ourssingular{} renderings. Our best results maximize the average transmittance of rays passing through the volume up to a target constant. Transmittance is the probability that light along ray $r$ is not absorbed by participating media when passing between point $t$ along the ray and the near plane at $t_n$~\eqref{eq:transmittance_continuous}.
We approximate the total transmittance along the ray as the joint probability of light passing through $N$ discrete segments of the ray according to Eq.~\eqref{eq:transmittance_discrete}.
Then, we define the following transmittance loss:
\begin{align}
    \Ltr &= -\min\!\left(\tau, \operatorname{mean}\!\left(T(\theta, \pose{})\right)\right)\\
    \Ltotal &= \Lclip + \lambda \Ltr
    \label{eq:additive_loss}
\end{align}
This encourages a \ourssingular{} to increase average transmittance up to a target transparency $\tau$. We use $\tau = 88\%$ in experiments.
$\tau$ is annealed in from $\tau=40\%$ over 500 iterations to smoothly introduce transparency, which improves scene geometry and is essential to prevent completely transparent scenes. Scaling $1-\tau\propto f^2 / d^2$ preserves object cross sectional area for different focal and object distances.

When the rendering is alpha-composited with a simple white or black background during training, we find that the average transmittance approaches $\tau$, but the scene is diffuse as the optimization populates the background. Augmenting the scene with random background images leads to coherent objects. \ours{} use  Gaussian noise, checkerboard patterns and the random Fourier textures from~\cite{mordvintsev2018differentiable} as backgrounds. These are smoothed with a Gaussian blur with randomly sampled standard deviation. Background augmentations and a rendering during training are shown in Figure~\ref{fig:bgaugs}.

\begin{figure}[t]
    \centering
    \ifarxiv
    \includegraphics[width=\linewidth]{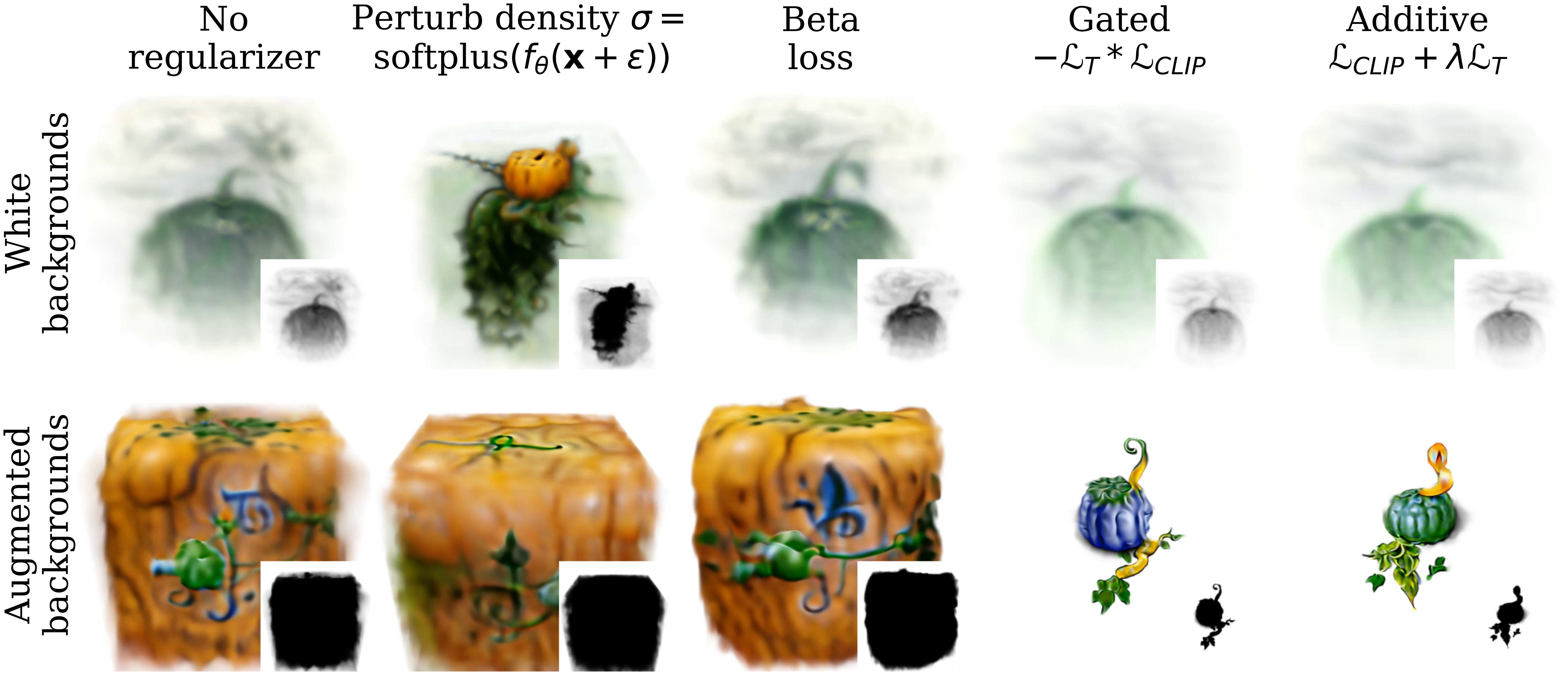}
    \else
    \includegraphics[width=\linewidth]{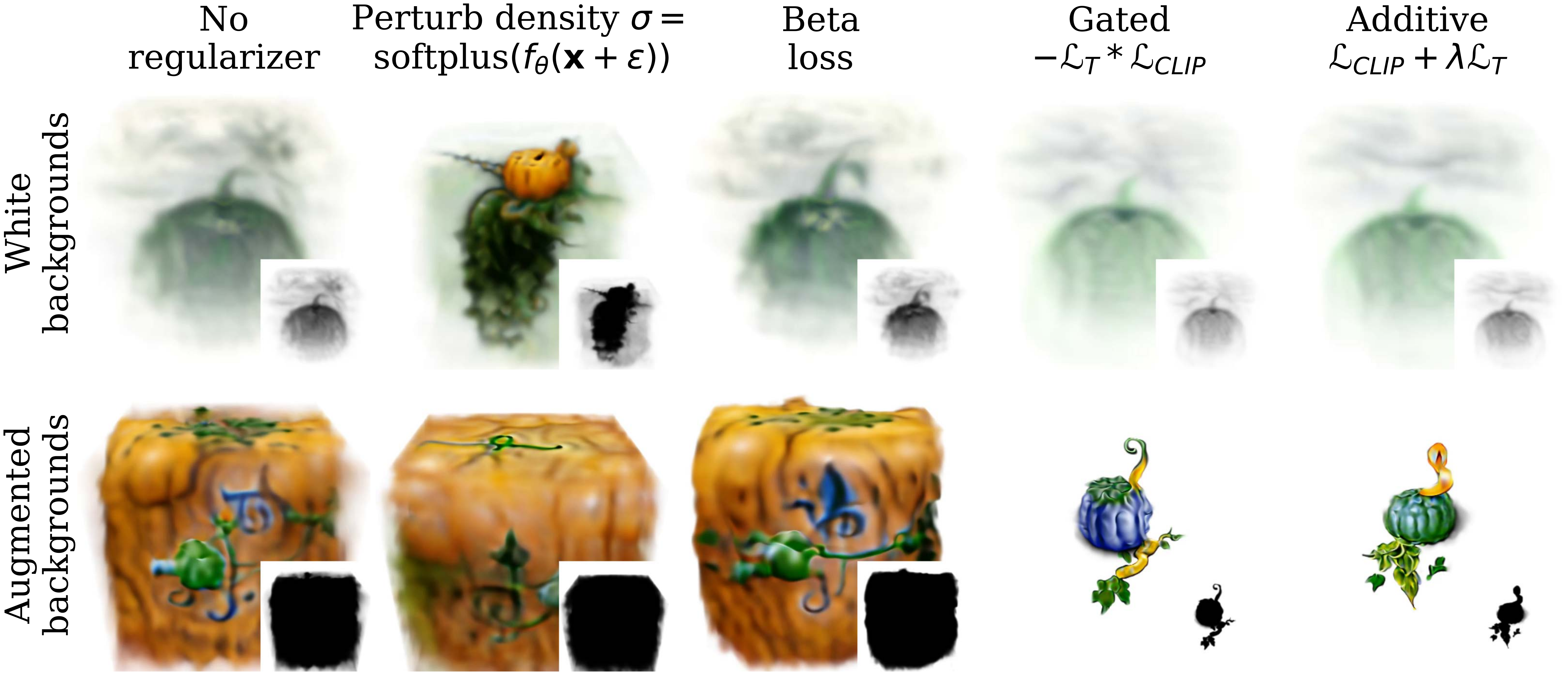}
    \fi
    \caption{Our transmittance losses and background augmentations are complementary. \textbf{Top:} Without background augmentations, priors on transmittance (right three columns) do not remove low-density structures. NeRF's density perturbations improve coherence, but cloudy artifacts remain. \textbf{Bottom:} When the object is alpha composited with random backgrounds during training, CLIP fills the scene with opaque material to conceal the background. However, gated and our simplified additive transmittance regularizers both limit the opacity of the volume successfully and lead to a sharper object. Inset panels depict transmittance. \textbf{Prompt:} \textit{``an illustration of a pumpkin on the vine.''}}
    \label{fig:bgvstransmittanceloss}
\end{figure}

We qualitatively compare \eqref{eq:additive_loss} to baseline sparsity regularizers in Figure~\ref{fig:bgvstransmittanceloss}. Our loss is inspired by the multiplicative opacity gating used by~\cite{mordvintsev2018differentiable}. However, the gated loss has optimization challenges in practice due in part to its non-convexity. The simplified additive loss is more stable, and both are significantly sharper than prior approaches for sparsifying Neural Radiance Fields.

\subsection{Localizing objects and bounding scene}

When Neural Radiance Fields are trained to reconstruct images, scene contents will align with observations in a consistent fashion, such as the center of the scene in NeRF's Realistic Synthetic dataset~\cite{mildenhall2020nerf}. \ours{} can place density away from the center of the scene while still satisfying the CLIP loss as natural images in CLIP's training data will not always be centered. During training, we maintain an estimate of the 3D object's origin and shift rays accordingly. The origin is tracked via an exponential moving average of the center of mass of rendered density. 
To prevent objects from drifting too far, we bound the scene inside a cube by masking the density $\sigma_\theta$.

\subsection{Neural scene representation architecture}

The NeRF network architecture proposed in~\cite{mildenhall2020nerf} parameterizes scene density with a simple 8-layer MLP of constant width, and radiance with an additional two layers. We use a residual MLP architecture instead that introduces residual connections around every two dense layers. Within a residual block, we find it beneficial to introduce Layer Normalization at the beginning and increase the feature dimension in a bottleneck fashion. Layer Normalization improves optimization on challenging prompts. To mitigate vanishing gradient issues in highly transparent scenes, we replace ReLU activations with Swish~\cite{DBLP:journals/corr/abs-1710-05941} and rectify the predicted density $\sigma_\theta$ with a softplus function. Our MLP architecture uses 280\textsc{K} parameters per scene, while NeRF uses 494\textsc{K} parameters.

\section{Evaluation}
\label{sec:evaluation}

We evaluate the consistency of generated objects with their captions and the importance of scene representation, then show qualitative results and test whether \ours{} can generalize compositionally.
Ablations analyze regularizers, CLIP and camera poses. Finally, supplementary materials have further examples and videos.

\subsection{Experimental setup}

3D reconstruction methods are evaluated by comparing the learned geometry with a ground-truth reference model, \textit{e.g.} with Chamfer Distance. Novel view synthesis techniques like LLFF \cite{mildenhall2019llff} and NeRF do not have ground truth models, but compare renderings to pixel-aligned ground truth images from held-out poses with PSNR or LPIPS, a deep perceptual metric \cite{zhang2018perceptual}.

As we do not have access to diverse captioned 3D models or captioned multi-view data, \ours{} are challenging to evaluate with geometric and image reference-based metrics. Instead, we use the CLIP R-Precision metric~\cite{park2021benchmark} from the text-to-image generation literature to measure how well rendered images align with the true caption. In the context of text-to-image synthesis, R-Precision measures the fraction of generated images that a retrieval model associates with the caption used to generate it. We use a different CLIP model for learning the \ourssingular{} and computing the evaluation metric. As with NeRF evaluation, the image is rendered from a held-out pose. \ours{} are optimized with cameras at a $30^\circ$ angle of elevation and evaluated at $45^\circ$ elevation. For quantitative metrics, we render at resolution 168$^2$ during training as in~\cite{Jain_2021_ICCV}. For figures, we train with a 50$\%$ higher resolution of 252$^2$.

We collect an object-centric caption dataset with 153 captions as a subset of the Common Objects in Context (COCO) dataset~\cite{10.1007/978-3-319-10602-1_48} (see supplement for details). Object centric examples are those that have a single bounding box annotation and are filtered to exclude those captioned with certain phrases like ``extreme close up". COCO includes 5 captions per image, but only one is used for generation. Hyperparameters were manually tuned for perceptual quality on a set of 20-74 distinct captions from the evaluation set, and are shared across all other scenes. Additional dataset details and hyperparameters are included in the supplement.

\subsection{Analyzing retrieval metrics}
\label{sec:evaluation:main_retrieval}

In the absence of 3D training data, \ours{} use geometric priors
to constrain generation. To evaluate each proposed technique, we start from a simplified baseline Neural Radiance Field largely following~\cite{mildenhall2020nerf} and introduce the priors one-by-one. We generate two objects per COCO caption using different seeds, for a total of 306 objects. Objects are synthesized with 10$\textsc{K}$ iterations of CLIP ViT B/16 guided optimization of 168$\times$168 rendered images, bilinearly upsampled to the contrastive model's input resolution for computational efficiency. R-Precision is computed with CLIP ViT B/32~\cite{clip:DBLP:journals/corr/abs-2103-00020} and \lituu{} B/32~\cite{zhai2021lit} to measure the alignment of generations with the source caption.

Table~\ref{tab:main_results} reports results. \textbf{The most significant improvements come from sparsity, scene bounds and architecture.} As an oracle, the ground truth images associated with object-centric COCO captions have high R-Precision. The NeRF representation converges poorly and introduces aliasing and banding artifacts, in part from its use of axis-aligned positional encodings.

We instead combine mip-NeRF's integrated positional encodings with random Fourier features, which improves qualitative results and removes a bias toward axis-aligned structures. However, the effect on precision is neutral or negative. The transmittance loss $\Ltr$ in combination with background augmentations significantly improves retrieval precision $+$18\% and $+$15.6\%, while the transmittance loss is not sufficient on its own. This is qualitatively shown in Figure~\ref{fig:bgvstransmittanceloss}. Our MLP architecture with residual connections, normalization, bottleneck-style feature dimensions and smooth nonlinearities further improves the R-Precision $+$8\% and $+$2\%. Bounding the scene to a cube improves retrieval $+$13\% and $+$11\%. The additional bounds explicitly mask density $\sigma$ and concentrate samples along each ray.

We also scale up \ours{} by optimizing with an image-text model trained on a larger captioned dataset of 3.6B images from \cite{zhai2021lit}. We use a ViT B/32 model with image and text encoders trained from scratch. This corresponds to the uu configuration from~\cite{zhai2021lit}, following the CLIP training procedure to learn both encoders contrastively. The \lituu{} ViT encoder used in our experiments takes higher resolution 288$^2$ images while CLIP is trained with 224$^2$ inputs. Still, \lituu{} B/32 is more compute-efficient than CLIP B/16 due to the larger patch size in the first layer.

\lituu{} does not significantly help R-Precision when optimizing \ours{} with low resolution renderings, perhaps because the CLIP B/32 model used for evaluation is trained on the same dataset as the CLIP B/16 model in earlier rows. Optimizing for longer with higher resolution 252$^2$ renderings closes the gap. \lituu{} improves visual quality and sharpness (Appendix~\ref{sec:additional_qual}), suggesting that improvements in multimodal image-text models transfer to 3D generation.

\begin{table}
  \centering
  \begin{tabular}{@{}llcHc@{}}
    \toprule
    & \multirow{2}{5mm}{Method} & \multicolumn{3}{c}{R-Precision $\uparrow$} \\
    & & CLIP B/32 & cos sim & \lituu{} B/32 \\ \midrule
    \multirow{2}{10mm}{Baseline} & COCO GT images & 77.1$\pm$\small{3.4} & -- & \pct{75.1634}{3.5045115342364735} \\
    & \multirow{1}{25mm}{Simplified NeRF} & \multirow{1}{*}{\pct{31.3725}{2.6569}} & \multirow{1}{*}{\metric{0.247129}{0.002182}} & \multirow{1}{*}{\pct{10.7843}{1.7761}}  \\ \midrule
    \multirow{3}{10mm}{Positional encoding}
    & ~~+ mip-NeRF IPE & \pct{29.7386}{2.6174} & \metric{0.245839}{0.002061} & \pct{12.4183}{1.8884}  \\
    & ~~+ \multirow{2}{22mm}{Higher freq.\\Fourier features} & \pct{24.1830}{2.4518} & \metric{0.234449}{0.002382} & \pct{10.4575}{1.7522} \\ & \\ \midrule
    \multirow{3}{10mm}{Sparsity,\\augment} & ~~+ random crops & \pct{25.8170}{2.5059} & \metric{0.234050}{0.002723} & \pct{10.4575}{1.7522} \\
    & ~~+ transmittance loss & \pct{23.6842}{2.4424} & \metric{0.226267}{0.002703} & \;\;\pct{7.5658}{1.5192} \\
    & ~~+ background aug. & \pct{44.1176}{2.8431} & \metric{0.255496}{0.002690} & \pct{26.1438}{2.5161} \\ \midrule
    \multirow{3}{10mm}{Scene\\param.}
    & ~~+ MLP architecture & \pct{51.9608}{2.8608} & \metric{0.268862}{0.002037} & \pct{27.7778}{2.5647} \\
    & ~~+ scene bounds & \pct{65.3595}{2.7246} & \metric{0.280474}{0.002048} & \textbf{\pct{38.8889}{2.7914}} \\
    & ~~+ track origin & \pct{59.8039}{2.8074} & \todo{NaN} & \pct{34.6405}{2.7246} \\ \midrule
    \multirow{3}{10mm}{Scaling} & ~~+ \lituu{} ViT B/32 & \pct{59.4771}{2.8111} & \metric{0.270799}{0.002208} & -- \\ 
    & ~~+ \multirow{2}{22mm}{20$\textsc{K}$ iterations, 252$^2$ renders} & \textbf{\pct{68.3007}{2.6643}} &  & -- \\ \\ \bottomrule
  \end{tabular}
  \caption{\textbf{When used together, geometric priors improve caption retrieval precision.} We start with a simplified version of the NeRF scene representation and add in one prior at a time until all are used in conjunction. Captions are retrieved from rendered images of the generated objects at held-out camera poses using CLIP's ViT B/32. Objects are generated with $\Lclip$ guidance from the pre-trained CLIP ViT B/16 except in scaling experiments where we experiment with the higher-resolution \lituu{} B/32 model.}
  \label{tab:main_results}
\end{table}

\begin{figure*}[th]
  \centering
  \includegraphics[width=\linewidth]{figures/shapes.pdf}
  \includegraphics[width=\linewidth]{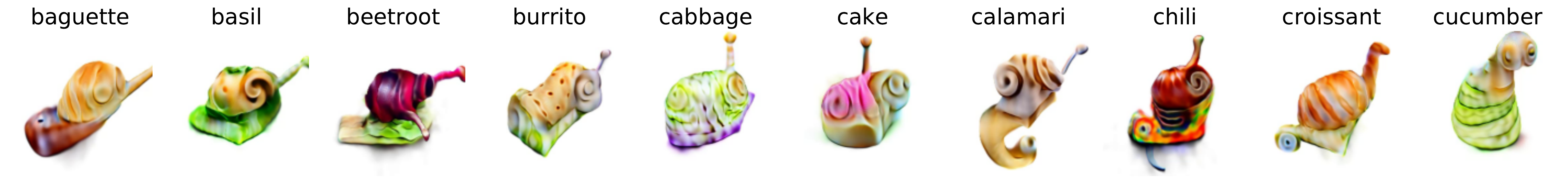}
  \caption{Compositional object generation. \ours{} allow users to express specific artistic styles via detailed captions.
  \textbf{Top two rows:} Similar to text-to-image experiments in~\cite{dalle:DBLP:journals/corr/abs-2102-12092}, we generate objects with the caption \textit{``\underline{armchair} in the shape of an \underline{avocado}. \underline{armchair} imitating \underline{avocado}.''} \textbf{Bottom:} Generations vary the texture of a single snail. Captions follow the template \textit{``a snail made of \underline{baguette}. a snail with the texture of \underline{baguette}''} Results are not cherry-picked.}
  \label{fig:compositional}
\end{figure*}

\subsection{Compositional generation}
\label{sec:evaluation:compositionality}

In Figure~\ref{fig:compositional}, we show non-cherrypicked generations that test the compositional generalization of \ours{} to fine-grained variations in captions taken from the website of \cite{dalle:DBLP:journals/corr/abs-2102-12092}. We independently vary the object generated and stylistic descriptors like shape and materials. 
DALL-E~\cite{dalle:DBLP:journals/corr/abs-2102-12092} also had a remarkable ability to combine concepts in prompts out of distribution, but was limited to 2D image synthesis. \ours{} produces compositions of concepts in 3D, and supports fine-grained variations in prompts across several categories of objects. Some geometric details are not realistic, however. 
For example, generated snails have eye stalks attached to their shell rather than body, and the generated green vase is blurry.

\subsection{Model ablations}

\begin{table}
  \centering
  \begin{tabular}{@{}lcc@{}}
    \toprule
    Method & Loss or parameterization & R-Prec. \\
    \midrule
    No regularizer & $\Lclip$ \eqref{eq:lclip} & \pctm{35.2941}{3.8762} \\
    Perturb $\sigma$~\cite{mildenhall2020nerf} & $\sigma=\operatorname{softplus}(f_\theta(\mathbf{x}) + \epsilon)$ & \pctm{47.7124}{4.0513} \\
    Beta prior~\cite{Lombardi:2019} & \eqref{eq:beta_prior} & \pctm{50.3268}{4.0554} \\
    Gated $T$~\cite{mordvintsev2018differentiable} & $-\operatorname{mean}(T(\theta, \pose{})) \cdot \Lclip$ & \pctm{34.6405}{3.8594} \\
    Clipped gated $T$ & $-\Ltr \cdot \Lclip$ \eqref{eq:gated} & \textbf{\pctm{62.0915}{3.9352}} \\
    Clipped additive $T$ & $\Lclip + \lambda \Ltr$ \eqref{eq:additive_loss} & \textbf{\pctm{62.0915}{3.9352}} \\
    \bottomrule
  \end{tabular}
  \caption{Ablating sparsity regularizers. Optimization is done for 10$\textsc{K}$ iterations at 168$^2$ resolution with \lituu{} ViT B/32 and background augmentation, and retrieval uses CLIP ViT B/32. For the purposes of ablation, we run one seed per caption (153 runs).}
  \label{tab:sparsity_ablation}
  \vspace{-2mm}
\end{table}

\begin{figure}[t]
    \centering
    \includegraphics[width=\linewidth]{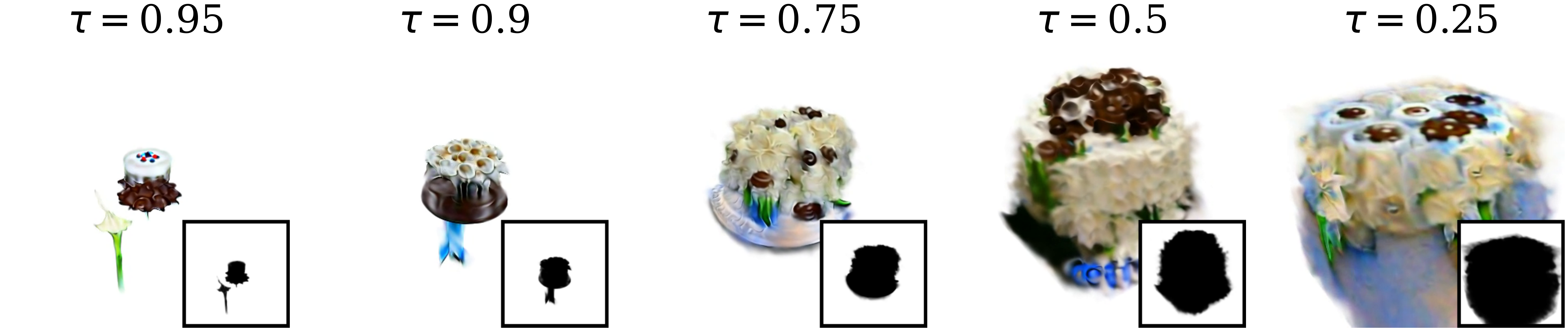}
    \caption{The target transmittance $\tau$ affects the size of generated objects. Inset panels depict transmittance. \textbf{Prompt from Object Centric COCO:} \textit{``A cake toped} [sic] \textit{with white frosting flowers with chocolate centers.''}}
    \label{fig:vary_transmittance}
\end{figure}

\noindent\textbf{Ablating sparsity regularizers}~~~~
While we regularize the mean transmittance, other sparsity losses are possible.
We compare unregularized \ours{}, perturbations to the density $\sigma$~\cite{mildenhall2020nerf}, regularization with a beta prior on transmittance~\cite{Lombardi:2019}, multiplicative gating versions of $\Ltr$ and our additive $\Ltr$ regularizer in Figure~\ref{fig:bgvstransmittanceloss}.
On real-world scenes, NeRF added Gaussian noise to network predictions of the density prior to rectification as a regularizer. This can encourage sharper boundary definitions as small densities will often be zeroed by the perturbation. The beta prior from Neural Volumes~\cite{Lombardi:2019} encourages rays to either pass through the volume or be completely occluded:
\begin{equation}
    \mathcal{L}_\mathrm{total}^\text{beta} = \Lclip + \lambda \cdot \operatorname{mean}\!\left(\log T(\theta, \pose{}) + \log (1-T(\theta, \pose{})) \right)
    \label{eq:beta_prior}
\end{equation}
The multiplicative loss is inspired by the opacity scaling of~\cite{mordvintsev2018differentiable} for feature visualization. We scale the CLIP loss by a clipped mean transmittance:
\begin{equation}
    \mathcal{L}_{\text{total}} = \min(\tau, \;\operatorname{mean}(T(\theta, \pose{}))) \cdot \Lclip
    \label{eq:gated}
\end{equation}
Table~\ref{tab:sparsity_ablation} compares the regularizers, showing that density perturbations and the beta prior improve R-Precision $+$12.4\% and $+$15\%, respectively. Scenes with clipped mean transmittance regularization best align with their captions, $+$26.8\% over the baseline.
The beta prior can fill scenes with opaque material even without background augmentations as it encourages both high and low transmittance. Multiplicative gating works well when clipped to a target and with background augmentations, but is also non-convex and sensitive to hyperparameters.
Figure~\ref{fig:vary_transmittance} shows the effect of varying the target transmittance $\tau$ with an additive loss.

\paragraph{Varying the image-text model}
We compare different image and text representations $h(\cdot), g(\cdot)$ used in $\Lclip$ \eqref{eq:lclip} and for retrieval metrics. Table~\ref{tab:model_ablation} shows the results. CLIP B/32, B/16 and \lituu{} B/32 all have high retrieval precision, indicating they can synthesize objects generally aligned with the provided captions. CLIP B/32 performs the best, outperforming the more compute intensive CLIP B/16 model. The architectures differ in the number of pixels encoded in each token supplied to the Transformer backbone, \textit{i.e.} the ViT patch size. A larger patch size may be sufficient due to the low resolution of renders: 168$^2$ cropped to 154$^2$, then upsampled to CLIP's input size of 224$^2$. Qualitatively, training with \lituu{} B/32 produced the most detailed geometry and textures, suggesting that open-set evaluation is challenging.

\begin{table}
  \centering
  \begin{tabular}{@{}lccc@{}}
    \toprule
    & \multicolumn{3}{c}{Retrieval model R-Precision} \\
    Optimized model & CLIP B/32 & CLIP B/16 & \lituu{} B/32 \\
    \midrule
    COCO GT & \pct{77.12418300653595}{3.406918943225708} & \pct{79.08496732026143}{3.29878833016494} & \pct{75.1634}{3.5045115342364735} \\  \midrule %
    CLIP B/32 \cite{clip:DBLP:journals/corr/abs-2103-00020} & \textit{(\pct{86.6013}{1.9505})} &  \pct{74.1830}{2.5059}  & \pct{42.8105}{2.8332} \\
    CLIP B/16 \cite{clip:DBLP:journals/corr/abs-2103-00020} & \pct{59.8039}{2.8074} & \textit{(\pct{93.4641}{1.4152})} & \pct{35.6209}{2.7420} \\  
    \lituu{} B/32   &  \pct{59.4771}{2.8111}  &  \pct{66.6667}{2.6993}  & \textit{(\pct{88.8889}{1.7995})} \\
    \bottomrule
  \end{tabular}
  \caption{The aligned image-text representation used to optimize \ours{} influences their quantitative validation R-Precision according to a held-out retrieval model. All contrastive models produce high retrieval precision, though qualitatively CLIP B/32 produced overly smooth and simplified objects. We optimize for 10$\textsc{K}$ iterations at 168$^2$ resolution. \textit{(Italicized)} metrics use the optimized model at a held-out pose and indicate \ours{} overfit.}
  \label{tab:model_ablation}
\end{table}

\begin{figure}
    \centering
    \includegraphics[width=\linewidth]{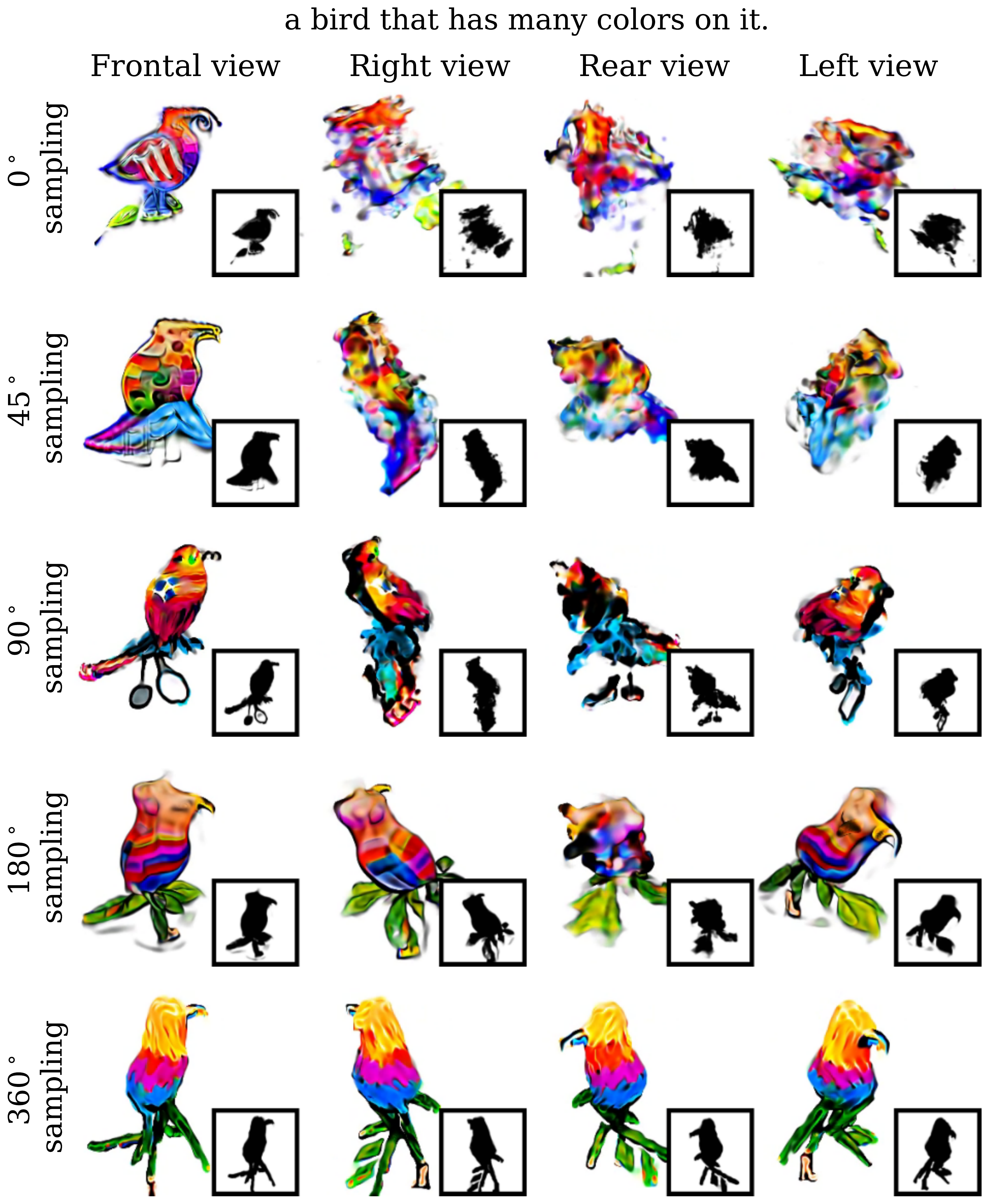}
    \caption{Training with diversely sampled camera poses improves generalization across views. In the top row, we sample camera azimuth from a single viewpoint. The rendered view from the same perspective (left column) is realistic, but the object structure is poor as seen from other angles. Qualitative results improve with larger sampling intervals, with the best results from 360$^\circ$ sampling.}
    \label{fig:vary_cameras}
    \vspace{-5mm}
\end{figure}

\vspace{2mm}\noindent\textbf{Varying optimized camera poses}~~~~
Each training iteration, \ours{} samples a camera pose $\pose$ to render the scene. In experiments, we used a full 360$^\circ$ sampling range for the camera's azimuth, and fixed the elevation. Figure~\ref{fig:vary_cameras} shows multiple views of a bird when optimizing with smaller azimuth ranges. In the left-most column, a view from the central azimuth (frontal) is shown, and is realistic for all training configurations. Views from more extreme angles (right, left, rear view columns) have artifacts when the \ourssingular{} is optimized with narrow azimuth ranges. Training with diverse cameras is important for viewpoint generalization.

\section{Discussion and limitations}

There are a number of limitations in \ours{}.
Generation requires iterative optimization, which can be expensive. 2\textsc{K}-20\textsc{K} iterations are sufficient for most objects, but more detail emerges when optimizing longer. Meta-learning~\cite{tancik2020meta} or amortization~\cite{Park_2019_CVPR} could speed up synthesis.

We use the same prompt at all perspectives. This can lead to repeated patterns on multiple sides of an object. The target caption could be varied across different camera poses.
Many of the prompts we tested involve multiple subjects, but we do not target complex scene generation~\cite{coyne2001wordseye, chang2014spatial, chang2015lexground, devries2021unconstrained} partly because CLIP poorly encodes spatial relations~\cite{liu2021learning,subramanian-etal-2022-reclip}. Scene layout could be handled in a post-processing step.

The image-text models we use to score renderings are not perfect even on ground truth training images, so improvements in image-text models may transfer to 3D generation. Our reliance on pre-trained models inherits their harmful biases. Identifying methods that can detect and remove these biases is an important direction if these methods are to be useful for larger-scale asset generation.

\section{Conclusion}
Our work has begun to tackle the difficult problem of object generation from text. By combining scalable multi-modal image-text models and multi-view consistent differentiable neural rendering with simple object priors, we are able to synthesize both geometry and color of 3D objects across a large variety of real-world text prompts. The language interface allows users to control the style and shape of the results, including materials and categories of objects, with easy-to-author prompts. We hope these methods will enable rapid asset creation for artists and multimedia applications.

\ifarxiv
\section*{Acknowledgements}
We thank Xiaohua Zhai, Lucas Beyer and Andreas Steiner for providing pre-trained models on the LiT 3.6B dataset, Paras Jain, Kevin Murphy, Matthew Tancik and Alireza Fathi for useful discussions and feedback on our work, and many colleagues at Google for building and supporting key infrastructure. Ajay Jain is supported in part by the NSF GRFP under Grant Number DGE 1752814.
\fi

{\small
\bibliographystyle{ieee_fullname}
\bibliography{main}
}

\clearpage
\appendix
\onecolumn

\begin{center}
\large \textbf{Zero-Shot Text-Guided Object Generation with \ours{}\\Supplementary Material}
\end{center}

\begin{wrapfigure}[14]{R}{0.6\textwidth}
    \vspace{-7mm}
    \centering
    \includegraphics[width=\linewidth]{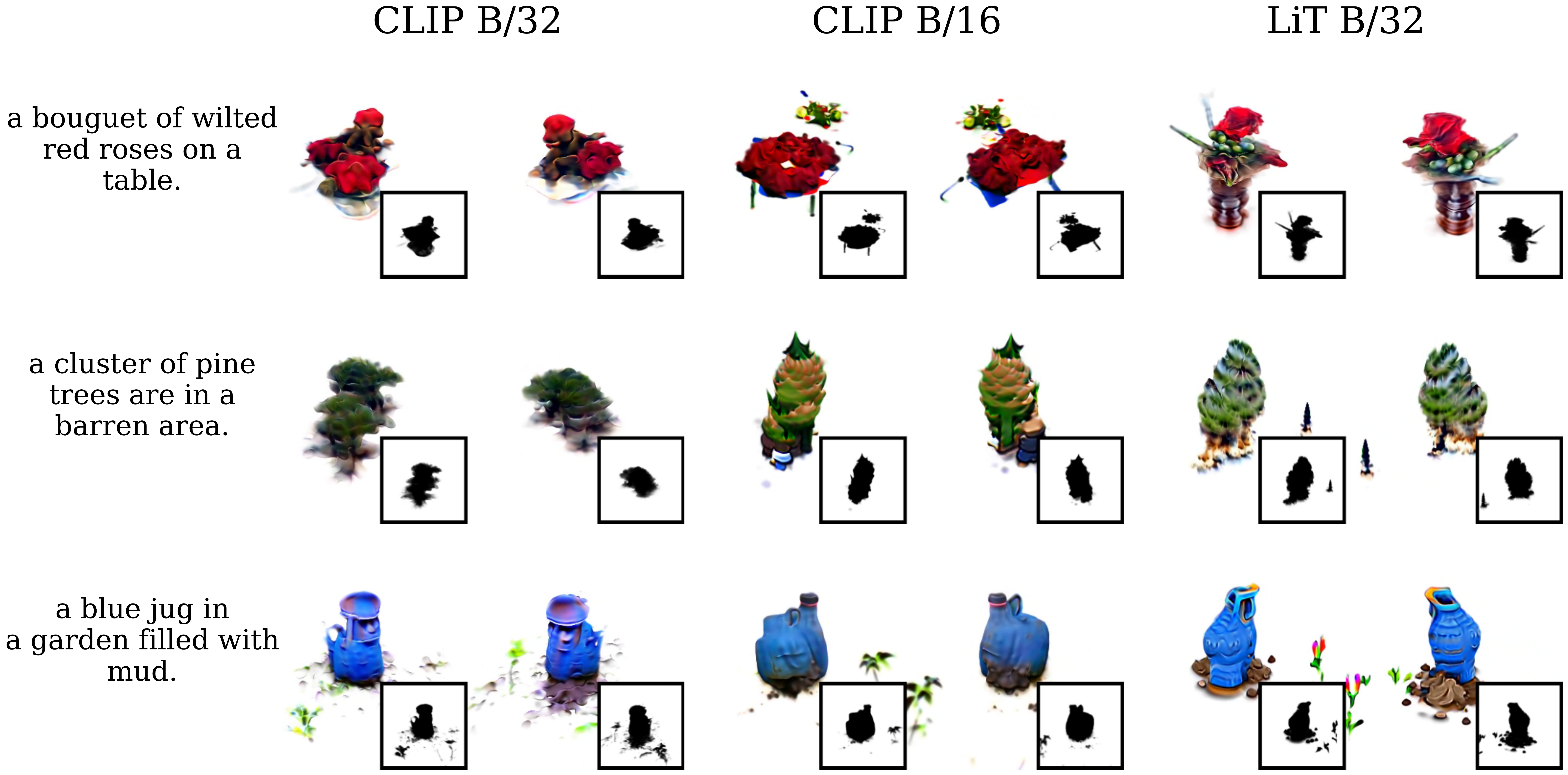}
    \caption{\textbf{Varying the image-text model} used for \ourssingular{} optimization.}
    \label{fig:vary_model}
\end{wrapfigure}
\section{Qualitative results and ablations}
\label{sec:additional_qual}
\textbf{An explanatory video with more qualitative results, code, an interactive Colab notebook, and object-centric prompts are available at {\small \url{https://ajayj.com/dreamfields}}}. The video includes 360$^\circ$ renderings where the camera orbits the object, as well as associated depth maps.

\begin{wrapfigure}{R}{0.5\textwidth}
    \vspace{-18mm}
    \centering
    \includegraphics[width=\linewidth]{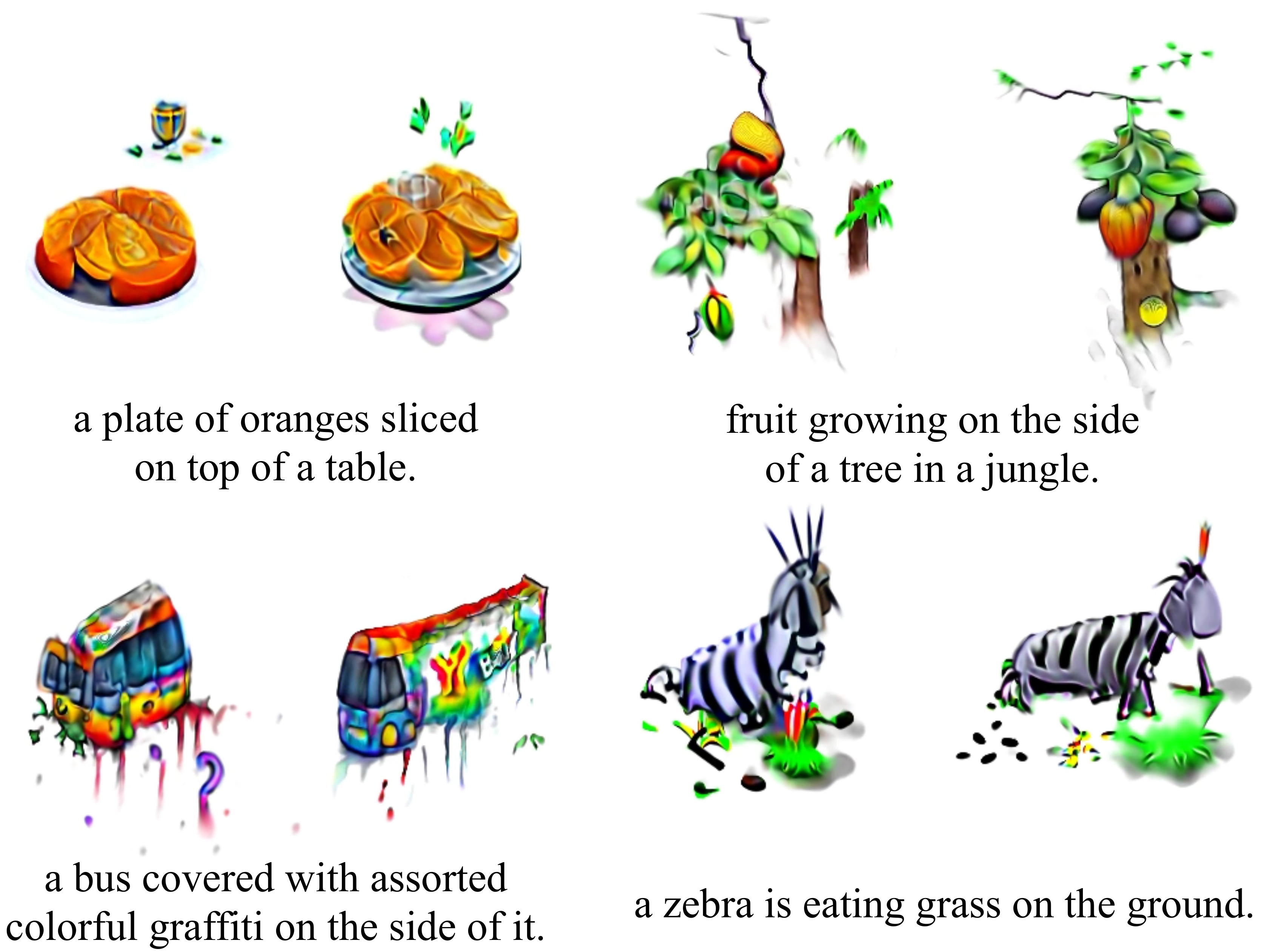}
    \caption{\textbf{Object diversity:} Objects vary with different seeds, effecting NeRF's weight initialization, camera sampling and augmentations.}
    \label{fig:multiple_seeds}
\end{wrapfigure}
\paragraph{Changing the image-text model} Figure~\ref{fig:vary_model} qualitatively compares generations using guidance from three different contrastive image-text models. \ours{} generated with \lituu{} B/32 are generally sharper than those with CLIP models, but all three can produce objects reflecting some aspects of the prompts. 

\paragraph{Diversity of synthesized objects} In creative applications, users often want to select between multiple synthesized results. \ours{} can synthesize multiple objects from the same prompt by changing the random seed before optimization. The seed changes the initialization of the NeRF weights, the camera pose sampled each iteration and the random background and crop augmentations. Figure~\ref{fig:multiple_seeds} shows the effect of changing the seed for four object-centric COCO prompts. Changing the seed changes scene shape and layout. For example, the bus on the left is compressed into a cubical shape, while the bus on the right is elongated. Colors and textures are often similar across the seeds, though can also vary. Changing the seed is another dimension of control in addition to prompt engineering.

\section{Object Centric COCO captions dataset}

Our Object Centric COCO dataset includes 153 test set prompts and 74 development set prompts. Several additional prompts are used for qualitative results, and are included in the main paper alongside figures. Captions are included along with code on the \href{https://ajayj.com/dreamfields}{project website}.

\newcommand{\myparagraph}[1]{\paragraph{#1} }

\section{Hyperparameters and training setup}

\myparagraph{Positional encoding} Our Fourier feature positional encodings use $L=8$ frequency levels, while novel view synthesis applications with image supervision commonly use $L=10$ to fit high-frequency details in photographs. Low-frequency ablations in Table~1 use $L=6$, which can improve convergence in the absence of our other geometric priors.

\myparagraph{Rendering}
Scenes are bounded to a cube with side length 2.
The camera is sampled at a fixed radius of 4 units from the center of the cubical scene bounds and an elevation of 30$^\circ$ above the equator. Near and far planes are set at $4\pm\sqrt{3}$ units from the camera based on the minimum and maximum possible distance to the corners of the cube. During training, we sample 192 points along each ray, spaced uniformly and jittered with uniform noise.
Rendered 168$^2$ views are cropped to 154$^2$ and upsampled to CLIP's input resolution for scenes where we compute qualitative metrics or 252$^2$ views are cropped to 224$^2$ for certain higher-quality visualizations. Crop sizes are selected to cover about 80\% of the image area. At test time, we sample 512 points along the rays and render at a higher resolution equal to CLIP's input size of 224$^2$ or LiT's input size of 288$^2$ for computing R-Precision and 400$^2$ for visualizations.

\begin{wrapfigure}{R}{0.3\textwidth}
\centering
\includegraphics[width=0.3\textwidth,trim={0 3mm 0 0},clip]{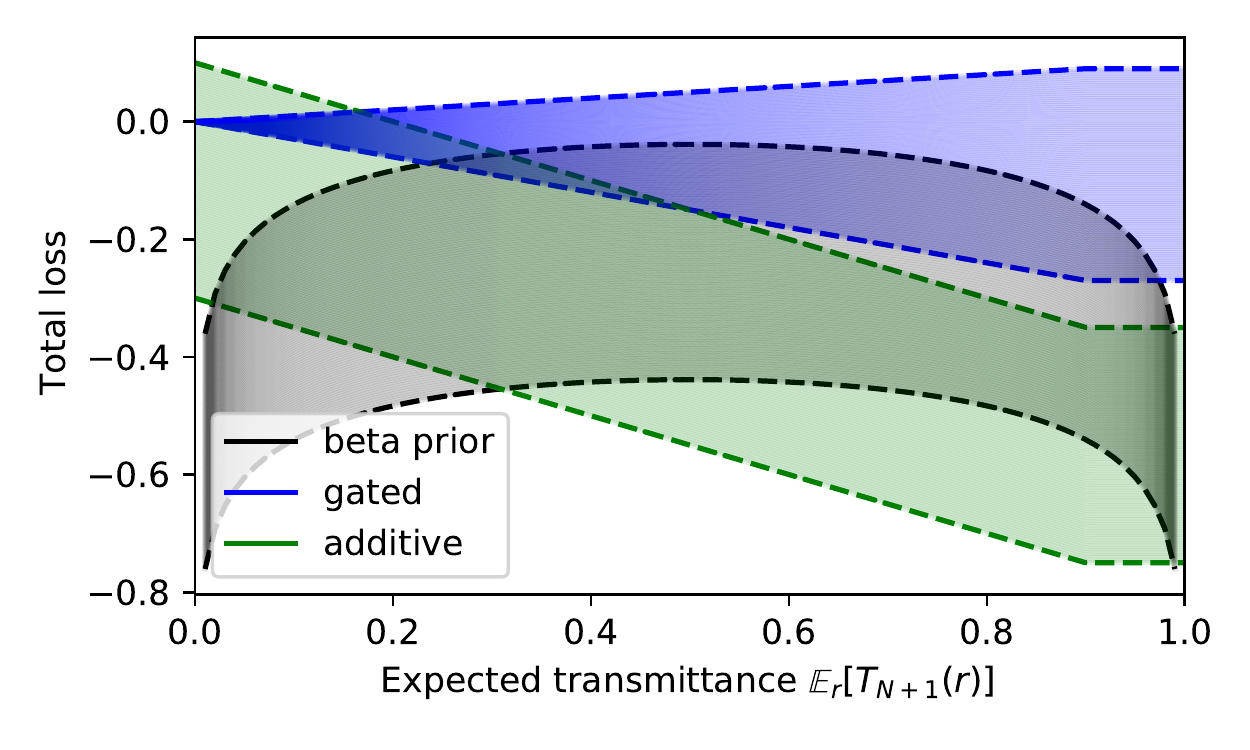}
\caption{\label{fig:sparsity_plot}Total loss with different sparsity regularizers and constant $\Lclip \in [0.1, -0.3]$. Upper and lower bounds of $\Lclip$ are shown with dashed lines. The additive loss is convex, and is always minimized by increasing transmittance.}
\end{wrapfigure}

\myparagraph{Optimization}
MLP parameters are initialized with the Flax~\cite{flax2020github} defaults: LeCun normal weights and zero bias for linear layers, and unit scaling and zero bias for layer normalization. The MLP is optimized with Adam with $\epsilon=10^{-5}$. Learning rate warms up exponentially from $10^{-5}$ to $10^{-4}$ over 1500 iterations, then is held constant. The camera origin is separately tracked with an exponential moving average with decay rate $0.999$ of the center of mass of rendered density. Figure~\ref{fig:sparsity_plot} visualizes different forms of the transmittance loss.

\begin{wrapfigure}[25]{R}{0.5\textwidth}
\centering
    \vspace{-1.5cm}
    \includegraphics[width=\linewidth]{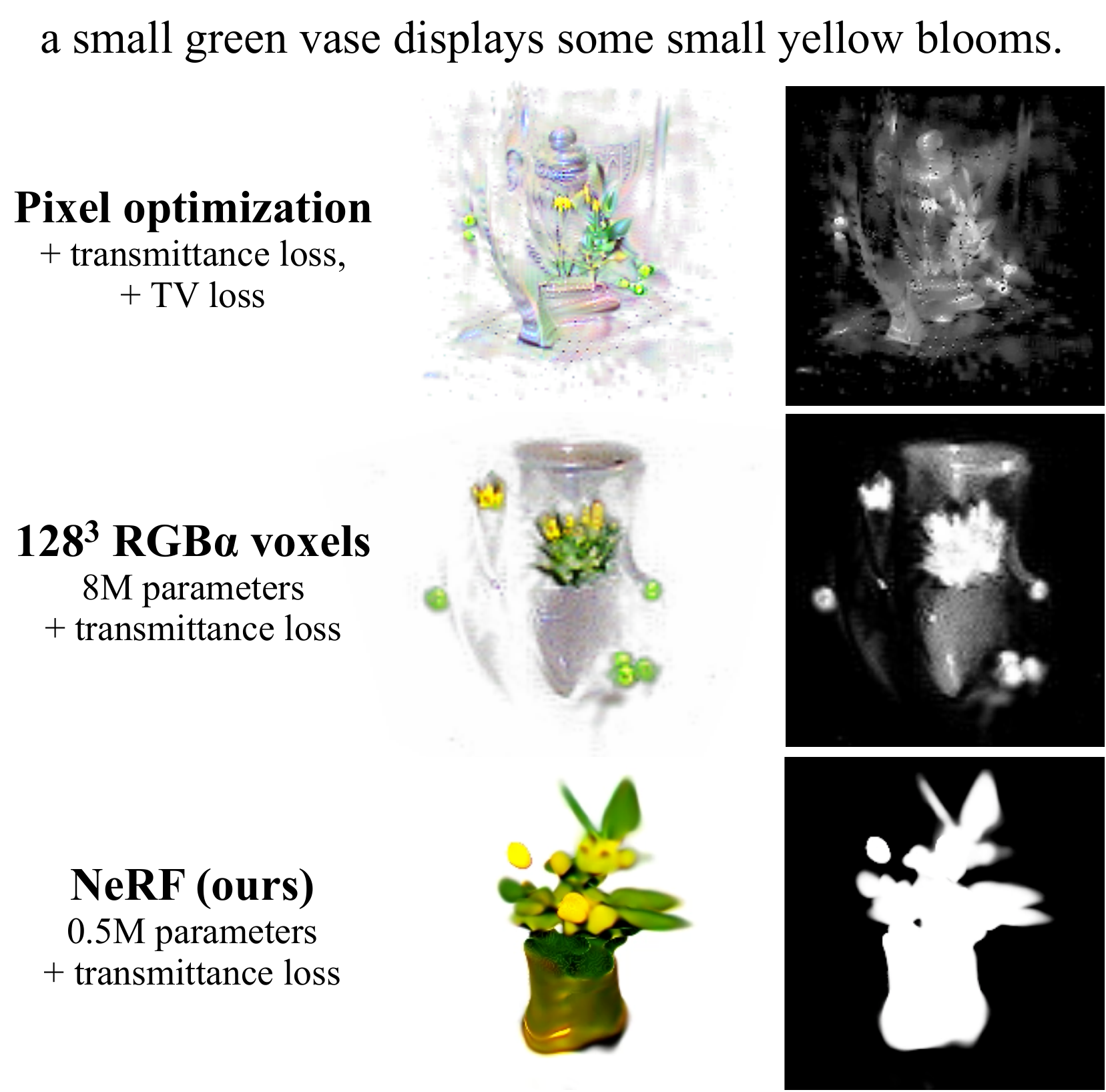}
    \caption{\textbf{Optimizing a Neural Radiance Field scene representation} (bottom) \textbf{leads to fewer artifacts} than optimizing an explicit single-view 2D image (top) or 3D voxel grid (middle), even when the explicit representations are regularized. Our MLP has 16$\times$ fewer parameters than the voxel grid, which may contribute to smoother, less noisy objects. We use CLIP B/16 for this experiment.}
    \label{fig:direct_optimization}
\end{wrapfigure}

\myparagraph{Hyperparameter selection}
Hyperparameters are manually tuned for visual quality on a development set of 74 object-centric COCO captions distinct from the test set reported in the paper. Most tuning is done on a smaller subset of 20 of the 74 captions, and hyperparameters are shared across all scenes.

\myparagraph{Hardware}
Optimization is done on 8 preemptible TPU cores, and 10$\textsc{K}$ iterations takes approximately 1 hour 12 minutes. This means each \ourssingular{} costs approximately \$3-4 to generate on Google Cloud, which is economical for applications. Training is bottlenecked by MLP inference and backpropagation during volumetric rendering, not CLIP.

\section{Pixel and voxel baselines}

We implemented 2D image optimization with a total variation loss and generative prior (CLIP Guided Diffusion), as well as a 3D voxel baselines to replace NeRF in Fig.~\ref{fig:direct_optimization}. All results for this ablation optimize CLIP ViT B/16.

The 2D image is an RGB$\alpha$ pixel grid, composited with random backgrounds during optimization similar to \ours{}.
Optimizing a single 2D RGB$\alpha$ image does not produce a multi-view consistent 3D object, so other viewpoints cannot be rendered. Even with transmittance and TV regularization, the resulting image is noisy.

The voxel grid stores 128$^3$ RGB and alpha values, 
interpolated trilinearly at ray sample points and composited without a neural network using the PyTorch3D library. Despite the transmittance loss, data augmentations and scene bounds, the voxel grid also has significant low-density artifacts. The voxel baseline has CLIP B/32 R-Precision 37.0\%{\footnotesize$\pm$3.9}, while NeRF has 59.8\%{\footnotesize$\pm$2.8} (Tables~\ref{tab:main_results}, \ref{tab:model_ablation}) with 16$\times$ fewer parameters, showing that the neural representation improves consistency with the input caption in a generalizable way. Using a hybrid representation with an explicit voxel grid followed by a smaller MLP head might improve computational efficiency of \ours{} without degrading quality.

\section{Signed distance field parameterization}

In early experiments, we learned scene density $\sigma$ with the VolSDF parameterization~\cite{yariv2021volume} $\sigma(\mathbf{x})=\alpha \Phi_\beta(-d_\Omega(\mathbf{x}))$ where $d_\Omega(\mathbf{x})$ is a signed distance function implicitly defining the object surface and $\Phi$ is the CDF of the Laplace distribution. This allows normal vector prediction with autodifferentiation and could improve the quality of the surface extracted from the radiance field. \ours{} successfully train with this alternate parameterization and produce visually compelling objects. The SDF Eikonal loss introduces an additional loss weight hyperparameter, which benefits from some tuning. Alternate 3D representations are an interesting avenue for future work.

\section{Impact of optimization time}

\begin{figure*}[t]
    \centering
    \includegraphics[width=\linewidth]{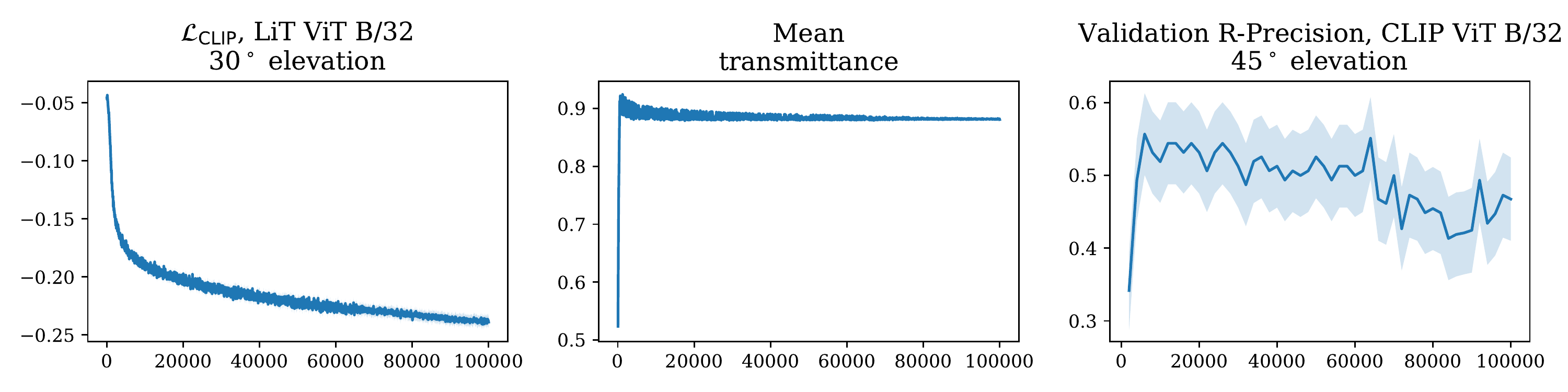}
    \caption{Long-run training and validation curves averaged over 79 hand-written prompts. Transmittance remains close to the target $\tau$ throughout training. \ours{} overfit to the image-text representations used for optimization, so in quantitative experiments, we stop training at 10$\textsc{K}$ iterations. The standard error of the mean is shaded.}
    \label{fig:overfitting}
\end{figure*}

\ours{} can overfit to the aligned image-text representation used for optimization. Figure~\ref{fig:overfitting} shows the training losses, $\Lclip$ and mean transmittance $\mathrm{mean}(T(\theta, \pose))$, as well as the validation R-Precision. Objects are generated with \lituu{} ViT B/32 guidance, and R-Precision is computed with a different contrastive image-text model, CLIP ViT B/32. Validation renderings are also done at a held-out elevation angle. Training loss continues to improve over long optimization trajectories, up to 10$\times$ longer than reported in the main paper. However, validation retrieval accuracy declines after 5-10$\textsc{K}$ iterations. The metrics are averaged over 79 different hand-written captions that test fine-grained variations in wording and prompt engineering.

\balance
Qualitatively, additional details and hyper-realistic effects are added over the course of long runs, shown in our supplementary video. Some details are not realistic, like floating text related to the typographic attacks identified in~\cite{goh2021multimodal}.

More augmentations may help further regularize the optimization. These include more aggressive 2D image augmentations such as smaller random crops, and more 3D data augmentations including varying focal length, varying distance from the subject and varying elevation. 3D data augmentations are supported by our approach.

\end{document}